%% file: main.tex
\documentclass{article}

\usepackage{iclr2027_conference,times}
\input{math_commands.tex}

\usepackage{hyperref}
\usepackage{url}
\usepackage{booktabs}
\usepackage{graphicx}
\usepackage{microtype}
\usepackage{tikz}
\input{figures/figure_icons.tex}
\usepackage{pgfplots}
\usepgfplotslibrary{fillbetween,groupplots}
\pgfplotsset{compat=1.16}
\input{figures/trajectory_styles.tex}
\usepackage{amsmath,amssymb,amsthm}
\usepackage[nameinlink,noabbrev]{cleveref}
\hypersetup{hidelinks,
  pdftitle={Task-Specific Agent Memory for Continual Learning},
  pdfauthor={Yezhou Cheng, Runjia Du, Zeming Liu, Qibai Chen, Hang Lyu, Yilan Wei, Yankai Zeng, Bojun Lin}}

\newtheorem{proposition}{Proposition}
\newcommand{\method}{\textsc{ORC}}
\newcommand{\bench}{\textsc{ProcStream-RSI}}

\title{Task-Specific Agent Memory\\
for Continual Learning}

\author{%
\begin{tabular}{c}
\href{https://openreview.net/profile?id=\%7EYezhou_Cheng1}{Yezhou Cheng}$^{1}$ \quad
\href{https://openreview.net/profile?id=\%7ERunjia_Du1}{Runjia Du}$^{2}$ \quad
\href{https://openreview.net/profile?id=\%7EZeming_Liu5}{Zeming Liu}$^{3}$ \quad
\href{https://openreview.net/profile?id=\%7EQibai_Chen1}{Qibai Chen}$^{4}$ \\
\href{https://openreview.net/profile?id=\%7EHang_Lyu2}{Hang Lyu}$^{3}$ \quad
\href{https://openreview.net/profile?id=\%7EYilan_Wei1}{Yilan Wei}$^{5}$ \quad
\href{https://openreview.net/profile?id=\%7EYankai_Zeng1}{Yankai Zeng}$^{6}$ \quad
\href{https://openreview.net/profile?id=\%7EBojun_Lin1}{Bojun Lin}$^{7}$ \\[0.4em]
\normalfont $^{1}$\,\href{https://www.wisc.edu/}{University of Wisconsin--Madison} \quad
$^{2}$\,\href{https://www.purdue.edu/}{Purdue University} \quad
$^{3}$\,\href{https://www.brown.edu/}{Brown University} \\
\normalfont $^{4}$\,\href{https://www.cmu.edu/}{Carnegie Mellon University} \enspace
$^{5}$\,\href{https://u.northwestern.edu}{Northwestern University} \enspace
$^{6}$\,\href{https://independent-researcher.org}{Independent} \enspace
$^{7}$\,\href{https://www.pinterest.com}{Pinterest, Inc.}
\end{tabular}%
}
\iclrfinalcopy

\begin{document}
\raggedbottom
\maketitle
\lhead{Preprint}

\begin{abstract}
Language-model agents can learn continually by storing useful instructions in memory and
reusing them as new tasks arrive, without changing model weights. This requires both learning
from new feedback and retaining earlier gains. However, an instruction learned from one task
family can hurt other families when the agent uses the same memory for every task. We introduce
task-specific agent memory: proposed skills are checked with execution tests, stored by task
family, and retrieved for tasks in that family. We evaluate this design on a 12-round code-repair
benchmark with recurring task families, tracking performance throughout learning. Holding
proposals, acceptance decisions, and task completions fixed, task-specific retrieval raises
average performance from 0.713 to 0.816. Across 27 paired randomized-order streams with
Qwen3-Coder-Next, it improves average performance by 0.063 over global memory and by 0.055
over an agent that never updates its memory, accumulating 63 skills with no observed harmful
acceptance. Capacity-matched retrieval and cumulative one-shot controls further support the
value of task-specific retrieval and sequential skill accumulation. Our contribution is to
identify cross-family interference and show how task-specific agent memory supports continual
learning through skill accumulation and retention.
\end{abstract}

\section{Introduction}

Large language models (LLMs) are increasingly used as agents for tasks such as code generation
and program repair. These agents often solve related tasks over multiple interactions. A coding agent, for example,
may repair programs involving boundary conditions, ordering, or normalization, and later return
to the same kinds of problems. The agent can use feedback from these tasks to revise a natural-language skill, store it in
persistent memory, and retrieve it for subsequent tasks.
Verbal feedback, iterative refinement, and automated agent design enable agents to improve
their instructions and workflows without updating model parameters
\citep{shinn2023reflexion,madaan2023selfrefine,hu2024adas}. Self-referential systems extend this
process by making edits persistent \citep{yin2024godelagent,zhang2025dgm}. In a deployed agent,
the objective is to learn from new feedback while retaining useful skills as different task
families recur. We study this form of \emph{continual learning through agent memory}: the
agent updates persistent natural-language skills while the underlying model remains frozen.

To preserve these gains, the agent needs to decide which tasks should use each skill. A repair
rule learned from boundary-condition examples may later enter prompts for ordering or
normalization tasks through a shared global memory. The rule has evidence from its source
family; its effects on other families are a separate question. An update gate can test the
current family and replay earlier families, while a globally retrieved skill also reaches
families that have not yet appeared. We call losses on other families caused by this exposure
\emph{cross-family interference}. Controlling retrieval gives the agent a way to retain the
local gain while managing where the new instruction is used.

Existing work shows that agents can improve their prompts, tools, and code. Persistent
deployment adds a complementary evaluation goal: tracking how each accepted update affects
performance as tasks recur. We therefore measure both gains on the current family and changes
on other families throughout the deployed sequence. Prior studies of self-correction
\citep{huang2023cannot}, actor--judge feedback \citep{pan2024rewardhacking}, and repeated
adaptation \citep{wang2026compound} motivate grounded tests, matched-compute controls, and
evaluation over time. We therefore ask: \emph{how can an agent retain a useful skill and
retrieve it for the tasks where that skill has been validated?}

We separate two decisions: whether to accept a skill update and which tasks should receive it.
Orthogonal Regression Control (\method) addresses the first through paired execution tests of
the current and proposed skills, combining public examples, private tests, and metamorphic
checks with historical-family regression checks. Task-specific retrieval addresses the second:
an accepted candidate is stored in its originating family's slot and supplied only when that
family recurs; a family without an accepted skill receives the initial skill. Both the actor and
editor remain frozen, and each task receives a single 360-character skill, so task-specific retrieval changes
which instruction is active without increasing the per-task skill budget. This design preserves the validation procedure while restricting each accepted skill to its
originating task family. In the experiments and figures, \emph{Scoped-\method{}}
denotes task-specific memory with the ORC gate. \Cref{fig:agent-memory-mechanism} shows how
an accepted skill enters memory and is reused when its family recurs.

\begin{figure}[!t]
    \centering
    \input{figures/agent_memory_mechanism.tex}
    \caption{Task-specific agent memory. ORC compares both skills on paired probes, requiring
    a current-family private-test gain and checking for regressions. The example boundary task
    reads a 360-character skill from slot $f$. Slot $g$ retains the ordering skill.}
    \label{fig:agent-memory-mechanism}
\end{figure}

We evaluate this approach on \bench, a procedural 12-round code-repair stream with recurring
task families. At each round, the actor repairs discovery programs, the editor proposes a skill
revision from visible feedback, and the gate either accepts the revision or retains the current skill.
Separate hidden checkpoints measure performance across all families throughout deployment,
including families not yet encountered in the update stream. We first isolate retrieval scope
using an archived-output intervention that holds proposals, gate decisions, task programs, and
scores fixed. We then run paired full-length streams in randomized family orders, allowing
subsequent proposals and gate decisions to evolve under each memory design.

The experiments connect a controlled mechanism to the performance of a complete adaptive agent.
The fixed-output intervention raises mean hidden trajectory utility from 0.713 with global memory
to 0.816 with task-specific memory, compared with 0.775 for the static agent. Across 27 randomized-order
streams, Scoped-\method{} improves trajectory utility by 0.063 over Global-\method{} and 0.055
over Static, accumulating 63 family skills with no observed harmful acceptance. Under matched
memory capacity, family slots improve trajectory utility by 0.026 over top-1 lexical retrieval.
Sequential updates also improve final-bank utility by 0.041 over writing skills once from
accumulated feedback. The gate comparisons explain a separate design choice: more permissive
selection can achieve higher average utility while allowing immediate losses. These findings
make retrieval scope, skill accumulation, and update selection explicit parts of memory design.

\begin{samepage}
Our contributions are:
\begin{enumerate}
    \item \textbf{A benchmark for continual learning through agent memory.} We introduce \bench,
    a 12-round code-repair stream with recurring task families and fixed hidden checkpoints.
    It measures performance throughout deployment, including gains on the current family and
    effects on other families.
    \item \textbf{A controlled identification of cross-family interference.} Holding proposals,
    gate decisions, and task completions fixed, we show that retrieval scope alone changes
    whether a locally useful skill improves or harms overall performance.
    \item \textbf{Task-specific agent memory and a systematic evaluation.} We combine
    execution-based update selection with family-specific skill retrieval. Paired experiments
    compare no updating, global memory, capacity-matched retrieval, simpler gates, learned
    routing, and cumulative one-shot writing. They establish gains on the Qwen task streams
    and reveal how routing and selection affect skill accumulation. A full-pipeline second-
    backbone experiment examines how the behavior changes with the actor and editor.
\end{enumerate}
\end{samepage}

Continual learning here takes place in the agent's persistent memory. Skills learned from
successive task batches remain available for future retrieval, and hidden checkpoints measure
both new gains and retention. In the Qwen streams, all 63 accepted skills create family slots:
the observed learning process accumulates skills across families. The cumulative one-shot
comparison tests the benefit of learning and using these skills as the sequence unfolds.

\section{Related Work}

Our work connects learning from feedback, persistent agent updates, and memory retrieval.
We organize the discussion around how agents acquire skills, validate changes, and reuse
experience on later tasks.

\paragraph{Self-correction and self-generated learning.}
Inference-time refinement and verbal memory can improve outputs without weight updates
\citep{madaan2023selfrefine,shinn2023reflexion}, whereas self-training changes weights using
known answers, filters, comparisons, or rewards \citep{zelikman2022star,singh2023restem,
yuan2024selfrewarding}. In both cases, the system needs a reliable way to select useful changes from generated candidates.
Studies of iterative feedback and selection examine how evaluation quality affects learning
and retention \citep{huang2023cannot,herel2024collapse,song2024mindgap,shafayat2025selftrain}.
We study skills stored as text in the prompt. Program execution provides feedback for deciding
whether to accept a proposed skill update.

\paragraph{Self-improving agents.}
Automated design searches over code, prompts, tools, and workflows \citep{hu2024adas}.
G\"odel Agent makes the task agent self-referential \citep{yin2024godelagent}; Darwin G\"odel
Machine retains an open-ended archive of self-edited coding agents \citep{zhang2025dgm}; and
later systems use two-timescale skills or comparative lineages
\citep{wang2026metaskill,liu2026mendel}. These works show that frozen models can improve the code, prompts, and tools used by an agent.
Our evaluation follows the skill used after each accepted update as different families recur.
A closely related continual evaluation studies transfer and second-phase improvement with
regression-aware optimization \citep{wang2026compound}. We examine the complementary role
of retrieval: how the tasks receiving a skill affect the value of a validated update.

\paragraph{Evaluator error dependence.}
Self-rewarding and meta-rewarding improve actors and judges together
\citep{yuan2024selfrewarding,wu2024metarewarding}, and evaluator co-evolution is made explicit
by the Red Queen G\"odel Machine \citep{iacob2026redqueen}. Actor--judge studies also examine shared feedback errors without gradient updates
\citep{pan2024rewardhacking}. Memory Reward
Inflation argues that corrective evidence must both track truth and have sufficiently distinct
errors \citep{asadolahi2026memory}. We use program execution and metamorphic checks in addition to the tests visible to the actor.
Separate hidden tests distinguish two problems: accepting an update that fails on the tested
family, and using a locally useful update on other families where it causes regressions.

\paragraph{Continual agent memory.}
AgentMemoryBench evaluates learning, retention, transfer, and repair in procedural and personal
memory. Its MEMs framework routes between specialized stores \citep{ma2026benchmarking}.
Synapse retrieves abstracted trajectories for prompting \citep{zheng2024synapse}, and Agent
Workflow Memory retrieves reusable workflow text \citep{wang2025agentworkflow}. Our
fixed-output intervention tests whether the same locally validated skill interferes with
performance when it reaches other families. The workflow-memory comparison adapts
relevance-based skill retrieval to the same task, editor, gate, and resource limits.

\paragraph{Code evaluation.}
EvalPlus expands correctness testing for generated code \citep{liu2023evalplus}; BigCodeBench
broadens instructions and library use \citep{zhuo2024bigcodebench}. Our procedural stream
supplies generated instances and held-out cases while keeping task semantics familiar and fixed;
it measures adaptation to recurring task requirements. A fixed HumanEval+ sample tests whether
the learned skills also help on an external benchmark.

\section{Continual Learning Through Agent Memory}

The agent learns continually by alternating between solving tasks and updating its memory. The actor
uses the skill to repair programs; the editor uses feedback from those repairs to propose a
revision; and the gate decides whether to keep it. Both actor and editor use frozen model
weights. We call tasks used to produce feedback \emph{discovery tasks}, and separate tasks used
to compare the current and proposed skills \emph{probe tasks}.

Let $\pi_\theta$ denote the frozen actor and $K_t$ its stored skill at round $t$. The actor
receives discovery tasks $D_t$, each containing a specification, buggy program, and visible
examples. It repairs the programs using $K_t$. The editor then uses these repairs and visible
failures to propose a revised skill $K'_t$. The gate $G_t$ compares both skills on the same
probe tasks. It keeps the revision if the candidate passes and otherwise retains $K_t$:
\begin{equation}
K_{t+1}=\begin{cases}K'_t,&G_t(K'_t,K_t)=1,\\K_t,&\text{otherwise.}\end{cases}
\end{equation}
The editor instruction is fixed. Any proposed changes to the editing instructions (meta-skills)
are ignored in confirmatory runs; only the task skill $K_t$ is carried into later rounds.
Repaired programs are task outputs and do not modify the agent's own code.

After each decision, we score the active skill on a fixed set of checkpoint tasks balanced
across families. These hidden scores never enter the update process. Let $U_t$ be the score
after round $t$. Our main measure is mean trajectory utility,
$\mathrm{MTU}=\frac{1}{T+1}\sum_{t=0}^{T}U_t$: average checkpoint performance over the initial
state and all update rounds. It captures what the agent delivers throughout the sequence.

We also report the last checkpoint score (terminal utility), changes on previously introduced
families (backward transfer), worst-family retention, and performance on a separate final test
bank. Primary claims summarize the deployed sequence, including every update decision.

\subsection{Why accepting harmful updates causes regression}

A gate must balance the gains from useful updates against the losses from harmful ones.
The following calculation makes this trade-off explicit.

Suppose a proposal is beneficial with probability $\pi$, has mean gain $G>0$ if beneficial,
and mean loss $L>0$ otherwise. Let a gate accept beneficial and harmful proposals with
probabilities $\alpha$ and $\beta$.

\begin{proposition}
The expected deployed change is positive exactly when
\begin{equation}
    \pi\alpha G > (1-\pi)\beta L.
\end{equation}
Suppose the gate requires every test channel to pass. If the channels falsely accept harmful
proposals independently, with rates $\beta_j$, the combined false-acceptance rate is
$\prod_j\beta_j$. If they falsely accept exactly the same proposals, adding channels leaves
$\beta$ unchanged.
\end{proposition}

In plain terms, accepted improvements must outweigh accepted regressions. An additional test
can strengthen selection by detecting errors missed by the existing tests. The calculation
shows why the overlap between their errors matters. We compare each test channel against hidden execution results, measuring
its ability to rank useful updates (AUROC), accepted updates that do not improve hidden scores,
harmful acceptances, and correlations between channel errors.
This calculation describes the outcomes represented in the gate tests. Our scope experiments
separately measure what happens when the skill also reaches families outside those tests.

\subsection{Task-specific skill retrieval}

Task-specific memory keeps a separate skill for each task family. Instead of replacing a skill
used for all tasks, an accepted update replaces only the current family's skill. For a task
from family $f$, the actor receives
\begin{equation}
K_t(f)=\begin{cases}K^{(f)}_t,&\text{if an accepted slot for $f$ exists},\\
K_0,&\text{otherwise.}\end{cases}
\end{equation}
The retrieved skill remains exactly 360 characters: retrieval changes which skill is supplied,
not the prompt budget. Proposal generation and the gate are unchanged. In the round-0
comparison, Global- and Scoped-\method{} therefore use the same discovery outputs, candidate,
probe evidence, and acceptance decision; only the scope of deployment differs. This comparison isolates the effect of supplying a skill to other families while keeping the
editor and gate fixed.

\section{Orthogonal Regression Control}

The gate checks whether a proposed skill improves the current family without causing regressions
on previously seen families. It runs the current and proposed skills on the same probe tasks.
The gate uses three kinds of evidence: four public examples, four private exact-output tests absent from the
actor/editor prompt, and at least six metamorphic checks. These checks test expected relationships between program
outputs without requiring the correct answer for each input. We call these evidence sources test channels. For task $i$ and channel
$c\in\{\mathrm{pub},\mathrm{priv},\mathrm{meta}\}$, $d_{i,c}$ is the proposed skill's
pass rate minus the current skill's pass rate.

We group probes by task family, keeping the current round separate from earlier rounds even
when the same family returns. For each group and test channel, \method{} computes a one-sided
95\% paired-bootstrap lower bound $L_{g,c}$ on the candidate's gain. The gate uses this lower
bound to check that an apparent improvement is supported across the probe tasks. Each group
has at least four tasks. We enumerate all bootstrap samples when $n^n\le 50{,}000$ and
otherwise use 2,000 draws with a fixed seed.

The candidate must improve on the current family's private tests, stay within the regression
tolerance on its other checks, and satisfy the tolerance for every historical group and channel.
With $\delta_{\mathrm{new}}=0$ and $\delta_{\mathrm{reg}}=0.02$, it is accepted only if
\begin{align}
L_{\mathrm{current},\mathrm{pub}}&\ge-\delta_{\mathrm{reg}}, &
L_{\mathrm{current},\mathrm{priv}}&>\delta_{\mathrm{new}},\\
L_{\mathrm{current},\mathrm{meta}}&\ge-\delta_{\mathrm{reg}}, &
L_{g,c}&\ge-\delta_{\mathrm{reg}}\quad\forall g<t,\forall c,
\end{align}
and no private or metamorphic evaluation reports an unsafe program. The gate checks each condition separately, keeping improvements and regressions visible
in every test channel.
Rejected proposals are logged but never enter later actor or editor prompts.

\section{Experimental Design}

The experiments first isolate task-specific retrieval, then measure continual learning
through skill accumulation and retention over complete task sequences. Additional comparisons examine memory capacity, update selection, routing,
and the timing of skill writes. The following subsections define the tasks and controls
used to answer these questions.

\subsection{\bench}

The benchmark measures continual learning by tracking skill acquisition and retention as
task families appear and recur. Each family groups programs with a shared repair requirement.
\bench~v3.1 uses the following 12-round schedule
\begin{center}
\small boundary, rotation, ordering, normalization, boundary, rotation, nested, missing,
prefix, chunk, tie-breaking, ordering.
\end{center}
Returning families let us measure whether earlier skills remain useful. Each family receives
an arbitrary protocol identifier determined by the random seed.
Discovery public examples distinguish the family-specific repair rule,
whereas probe, checkpoint, and final public examples are neutral: the supplied buggy program
already passes them. Their private, metamorphic, and hidden cases exercise that rule. Importantly,
the required program behavior and family order are fixed in this experiment. Seeds change
tokens, names, program variants, and test cases, while preserving each family's repair rule. The benchmark therefore measures
adaptation to new instances of recurring, familiar repair requirements. Static's 0.775 mean trajectory utility is consistent with substantial prior competence.

Each round has four discovery and four probe tasks. Separately generated fixed banks contain
one checkpoint and two final tasks per family, for 123 tasks per stream. Exact-output public,
private, and hidden inputs are pairwise disjoint, and hidden cases are unavailable to the online
agent. Public task views remain unique after normalizing their role labels. We also check the task
manifests---the files listing the tasks and their evaluation cases---as described in the
reproducibility statement and supplement. \Cref{fig:experiment-protocol} shows when
the different task sets are used.

\begin{figure}[t]
    \centering
    \input{figures/experiment_protocol.tex}
    \caption{Task and test roles in the full 12-round protocol. Each update uses discovery
    feedback and a separate probe batch. Hidden utility is measured on the same checkpoint
    task bank before the first update and after every decision. A separate final bank evaluates
    the final memory state. Hidden scores are computed offline from archived completions.}
    \label{fig:experiment-protocol}
\end{figure}

\subsection{Methods and controls}

We run seven methods on matched streams to separate the effects of updating, selection, and
additional model calls:
\begin{enumerate}
    \item \textbf{Static}, which never proposes an update;
    \item \textbf{Frozen-compute}, which makes the same editor, replay, gate, and evaluation
    calls as \method~but never installs an update;
    \item \textbf{Latest-only}, which deploys every nonempty, well-formed proposal;
    \item \textbf{Self-judge}, whose actor estimates correctness of current and proposed
    programs and accepts the higher mean estimate;
    \item \textbf{Replay}, which accepts when average public-test performance improves on the
    current probes and does not decrease on historical probes;
    \item \textbf{\method}, the gate requiring all test channels to pass; and
    \item \textbf{Batch-\method}, which gives the editor accumulated discovery feedback but
    keeps the initial skill in use until the final proposal can be accepted.
\end{enumerate}
Batch-\method~provides the one-shot control for the eight-stream global experiment.
Frozen-compute tests the effect of additional calls. A hidden-gated oracle is used only for
code-path validation.

\subsection{Scope and randomized-entry extension}

We first isolate retrieval scope on the eight main streams. Every Global-\method{} run accepts
the same kind of round-0 boundary rule, so task-specific retrieval uses its archived completion on
boundary tasks and the archived Static completion elsewhere. The proposals, gate decisions, task
programs, and scores remain unchanged; only the task-to-skill assignment differs.

Next, a one-update experiment tests the retrieval comparison with each of the nine families
appearing first. The v4 generator
reorders the same family occurrences using a permutation fixed by the seed. Eighteen seeds give two
streams for each of the nine possible first families. Each stream contains one update round,
four discovery and four probe tasks, and checkpoint and final banks with two tasks per family.
Global- and Scoped-\method{} are paired within stream and must share their proposal and gate
decision. The primary comparison is Scoped minus Global hidden checkpoint utility after that
decision; secondary outcomes are final-bank utility, harmful accepted updates, and changes on
the current versus eight non-current families. We report stream-bootstrap 95\% intervals and a
two-sided paired sign-flip test.

\subsection{Model, runs, and outcomes}

The actor/editor is Qwen3-Coder-Next served by the pinned Parasail BF16 endpoint through
OpenRouter, with provider fallback disabled, temperature 0, a fixed stream seed, 700 actor
tokens, and 500 editor tokens. The model was selected on public smoke tasks for response and
format reliability before the v3 pilot. Actor skill text is padded to 360 characters and editor
input to 48,000 characters, making prompt budget independent of skill length. Batch-\method's
cumulative-evidence editor is instead padded to 64,000 characters on every round, and its larger
token cost is reported separately. Model weights are never trained.

After a two-seed, six-round difficulty pilot, all thresholds and prompt content were fixed. The
main study uses eight unseen seeds and all 12 rounds. The statistical unit is a complete stream, rather than an individual task or update.
Primary contrasts are \method--Replay mean trajectory utility, \method--Latest-only mean
trajectory utility, and
\method--Batch-\method~terminal checkpoint utility. We report paired raw seeds, bootstrap 95\%
intervals, two-sided paired sign-flip sensitivity tests, and Holm correction across these three
tests. The analysis covers these eight seeded streams; task-level results are descriptive. We use
provider-reported token counts to estimate the cost of running without shared cached responses.

The randomized-entry extension keeps the Qwen endpoint and actor/editor limits fixed.
Transfer diagnostics use a fixed HumanEval+ sample and a GPT-OSS-120B actor with frozen
final skills. \Cref{app:secondary-diagnostics} gives their protocols.

Finally, a prespecified full-length extension uses 27 seeds, exactly three for each possible first
family, and pairs Global- with Scoped-\method{} within every stream. Skills evolve for all 12
rounds and may accumulate multiple accepted slots. The primary outcome is mean hidden trajectory
utility. Secondary outcomes include final-checkpoint and final-bank utility, acceptance counts,
and harmful acceptances. Online Scoped-\method{} retrieves skills using the known family labels in the task manifest. The same-stream Static reference reuses audited initial-skill completions on
exactly the same checkpoint and final tasks. Separate retrieval tests hold the final memories fixed and vary which skill each task receives.

The follow-up uses the same 27 task manifests, model, prompts, and feedback permissions.
For both global and task-specific memory, we compare Latest-only, Replay, private-only, and full \method{}.
Additional task-specific-memory controls compare nine family slots with matched multiversion storage,
replace oracle slot routing with a frozen learned router, and delay all writes until
the last round. \Cref{app:reviewer-controls} specifies their matched budgets and retrieval rules.

We rerun Static, Global-\method{}, Scoped-\method{}, and matched Multiversion-\method{} on the same
27 streams with GPT-OSS-120B as actor and editor \citep{openai2025gptoss}. All four start from
the initial skill and run 12 rounds with the original prompts and feedback (\cref{app:second-backbone}).

\section{Results}

The results address three questions: does retrieval scope explain cross-family interference,
does task-specific memory improve performance over complete task streams, and which memory and
selection choices account for that improvement? We first isolate the mechanism, then compare
continual learning with task-specific memory with matched baselines and test routing and sequential accumulation. A harmful acceptance is an accepted update that lowers
the next hidden checkpoint score.

\IfFileExists{generated_results.tex}{\input{generated_results.tex}}{}

\IfFileExists{generated_scoped.tex}{\input{generated_scoped.tex}}{}

\begin{figure}[t]
    \centering
    \input{figures/family_interference.tex}
    \caption{Cross-family effects of the first accepted update on eight fixed-order streams.
    Within each stream, both retrieval policies use the same accepted boundary skill and
    archived completions.
    With task-specific retrieval, only boundary tasks receive the update and the other
    families retain their initial-skill scores. Points show descriptive means by family.}
    \label{fig:family-interference}
\end{figure}

\paragraph{Family-level gains reveal the need for task-specific retrieval.}
Static, Frozen-compute, and Batch-\method{} all score 0.775 in trajectory utility on the
eight fixed-order streams. The adaptive global baselines score 0.703--0.736
(\cref{tab:main-results,fig:trajectory}). Global-\method{} accepts only the first proposal
in every stream, before any historical family has appeared. Six of the eight deployments
reduce the next checkpoint. The change raises boundary-family utility by 0.398 and changes
missing, rotation, tie-breaking, and chunk utility by $-0.461$, $-0.234$, $-0.148$, and
$-0.117$. \Cref{fig:family-interference} shows how these effects change when the same
update is restricted to boundary tasks.

\paragraph{Task-specific retrieval preserves local gains across families.}
Changing only retrieval scope yields the central improvement (\cref{tab:scoped-retrieval}).
Using the same archived task completions, Scoped-\method{} attains mean trajectory utility 0.816,
compared with 0.713 for Global-\method{} and 0.775 for Static. Its paired advantage is 0.103 over
Global-\method{} (bootstrap 95\% interval $[0.052,0.151]$, sign-flip sensitivity $p=0.03125$)
and 0.041 over Static ($[0.022,0.056]$, $p=0.03125$). The same eight locally accepted updates
raise the immediate global checkpoint by 0.044 on average when retrieved only for boundary
tasks, and none is harmful; global deployment made six harmful. Task-specific retrieval avoids a mean round-0 interference loss of 0.112 while keeping the active
family's task completions unchanged. The
randomized-entry experiment below tests the same mechanism across every possible first family;
the full extension then tests whether task-specific retrieval supports repeated updating.

\IfFileExists{generated_randomized_entry_prose.tex}{\input{generated_randomized_entry_prose.tex}}{}

\input{generated_static_comparison.tex}

\paragraph{Task-specific memory improves performance throughout the task sequence.}
Across the 27 full streams, Scoped-\method{} improves mean trajectory utility by 0.055 over
Static ($[0.040,0.071]$) and by 0.063 over Global-\method{} ($[0.037,0.094]$).
\Cref{fig:randomized-full} shows the checkpoint trajectories. The gains
also remain at the end: final-bank improvements are 0.098 over Static ($[0.074,0.122]$)
and 0.118 over Global ($[0.076,0.168]$). Final-checkpoint gains are 0.107
($[0.079,0.134]$) and 0.111 ($[0.078,0.150]$), respectively. All three Global-minus-Static
intervals include zero.

The agent with task-specific memory builds up useful skills across families. It creates 63 family slots in total,
with multiple slots in 19/27 streams. These are first writes: none of the 17 returns to an
already populated slot produces an accepted replacement.

\begin{figure}[t]
    \centering
    \input{figures/randomized_trajectories.tex}
    \caption{Full randomized-order extension. (a) Mean hidden checkpoint trajectories across 27
    paired 12-round streams. (b) Retrospective sensitivity to falling back to the initial skill
    when routing misses. The leave-one-seed-out router uses task descriptions that share
    templates across seeds. Shading in both panels shows 95\% stream-bootstrap intervals.}
    \label{fig:randomized-full}
\end{figure}

\paragraph{Gate choice balances higher utility against immediate regressions.}
With task-specific memory, permissive gates achieve higher utility but accept some updates that lower
the next checkpoint. Scoped-Replay improves trajectory utility by 0.036 over Scoped-\method{}
($[0.020,0.053]$), and its final-bank mean is 0.953 versus 0.862. Of its 287 accepted updates,
11 lower the next checkpoint, affecting 11/27 streams. Scoped-Latest also has higher final-checkpoint and final-bank means, with 24 harmful acceptances in 20/27 streams.

Private tests reproduce all 324 Scoped-\method{} decisions and all three utility outcomes.
The private-test rule and full gate give the same measured selection outcome on these streams. With global memory, the pattern
is different: Replay and Latest have lower trajectory means than \method{}
(\cref{app:reviewer-controls}). Thus, retrieval scope changes the consequences of accepting
more updates, while the gate controls how often immediate losses occur.

\paragraph{Family slots outperform the matched lexical workflow store.}
With the same nine-entry capacity, metadata allowance, and 360-character active skill
limit, family slots gain 0.026 ($[0.012,0.043]$) in trajectory utility and 0.060
($[0.026,0.099]$) on the final bank. Multiversion retrieval incurs 20 nontarget-family
decreases across ten accepted deployments. Both methods use the same editor and gate; the workflow store retrieves one skill using
lexical similarity (\cref{app:reviewer-controls}).

\paragraph{Accurate routing helps preserve the benefits of task-specific memory.}
The online learned router achieves 0.810 trajectory utility, compared with 0.848 when retrieval
uses the known family labels. The paired gap is 0.039 ($[0.026,0.052]$). Fourteen of its
46 accepted updates lower performance on a family other than the current one.

A separate test holds the final memories fixed and deliberately retrieves the wrong skill.
Mean final-bank utility falls to 0.686, compared with 0.764 for the initial skill and 0.862
for the correct family skill. This separates the cost of an incorrect retrieval from simply
falling back to the initial skill. Read counts and routing rules appear in
\cref{app:reviewer-controls}.

\paragraph{Continual skill accumulation improves final performance over one-shot writing.}
Scoped-Cumulative waits until the final round to write skills from accumulated discovery
feedback. Under the same slot budget, sequential Scoped-\method{} improves the final-bank
score by 0.041 ($[0.023,0.062]$) and final-checkpoint utility by 0.051 ($[0.026,0.078]$).
The cumulative control accepts 34 skills at the last decision, compared with 63 first writes
across the sequential streams. This comparison supports learning and using family skills
as the sequence unfolds, under the tested feedback and selection protocol.
\Cref{fig:memory-controls} collects the paired effects for the retrieval and timing controls.

\begin{figure}[t]
    \centering
    \input{figures/memory_controls.tex}
    \caption{Paired effects of retrieval and update timing on 27 randomized Qwen streams.
    Dots are mean utility differences (first method minus its control). Bars are 95\% intervals
    from resampling whole paired streams. The lexical control matches store and active-skill
    capacity. The one-shot comparison reports its two prespecified final endpoints.}
    \label{fig:memory-controls}
\end{figure}

\paragraph{A second backbone tests adaptation from a stronger initial policy.}
Mean trajectory utility is 0.914 for Scoped- and Global-\method{}, 0.916 for
Multiversion-\method{}, and 0.909 for Static. All three paired comparisons with Scoped-\method{} have intervals that include zero
(Holm $p=1$). Scoped and multiversion each accept two updates, and global accepts one.
No harmful acceptance is observed (\cref{app:second-backbone}).

\section{Conclusion}

Task-specific agent memory supports continual learning by retaining skills from earlier
feedback and retrieving them for the corresponding tasks. Our fixed-output intervention identifies why this matters: a locally useful
repair rule can improve overall performance under task-specific retrieval while causing losses when
retrieved globally. On the 27 Qwen streams, task-specific agent memory improves performance over
both global memory and Static, and matched-capacity and one-shot controls support the value
of task-specific retrieval and sequential skill accumulation. The gate comparisons further
show how to distinguish average utility from immediate regression control. These results
provide an experimentally grounded design principle for memory-based continual learning:
validate each skill, match its use to the corresponding task family, and track acquisition
and retention throughout deployment. The second-backbone experiment and learned-routing controls show how initial competence,
update frequency, and task assignment shape this process.

\section{Limitations and Broader Impact}

We study two frozen backbones with different endpoints and generation allowances on nine generated
code-repair families. The learned router uses public I/O structure, and ORC uses known family labels
for evaluation groups. The workflow comparison covers lexical top-1 retrieval. The one-shot comparison evaluates
learning from feedback collected under the initial skill versus feedback collected as skills
evolve. Open-ended tasks and additional retrieval strategies are natural next settings for
evaluating the same design principle.

No harmful acceptance was observed among the 63 Qwen task-specific-memory updates across 27 streams. Updates
within a stream are dependent. A stream-level Wilson 95\% interval for observing at least one
harmful accepted update is $[0,0.125]$ for this cohort.

Generated programs run in a pinned, network-free Docker sandbox with least privilege, resource
limits, and abstract syntax tree (AST) validation. The study has no personal data or human subjects. We report full performance trajectories and costs to make the results and their experimental
setting explicit.

\clearpage
\bibliography{references}
\bibliographystyle{iclr2027_conference}

\section*{Reproducibility}

The supplement provides the materials needed to reproduce task generation, agent updates,
and evaluation: generator and evaluator code, task manifests and checksums, exact prompts,
seeds, endpoint settings, accepted and rejected skills, archived model responses, analysis
scripts, external task IDs, and dated protocols and deviations.

We keep update feedback separate from hidden evaluation. Online runners reject manifests that
contain hidden tests. The offline evaluator checks for missing cases, inconsistent manifests,
duplicate artifacts, incorrect data splits, and changed code hashes. Hidden scores are computed
from the archived programs without further model calls.

Incomplete runs are archived and restarted from round zero with a new run identifier; partial
runs never enter the analysis. Requests that time out after dispatch may still incur a charge,
so they are recorded conservatively and are not automatically retried. A shared transactional
ledger tracks spending and cache reuse. Method-level cost comparisons reconstruct uncached
endpoint use from provider-reported usage. The randomized Qwen extensions use the same
accounting rules and endpoint as the main study. The supplement includes paired run records,
excluded attempts, and the unit and sandbox tests. The reviewer follow-up incurred
\$4.671 including conservative charges for ambiguous dispatches. The Qwen-centered study
spent \$16.234 under the original \$17 cap. The full GPT-OSS-120B rerun adds \$1.983 in
settled charges, bringing cumulative spend to \$18.217 under the authorized \$30 cap.

\section*{AI Use Statement}

Generative AI tools were used to identify and summarize literature, formulate hypotheses,
design the benchmark and experiments, implement and test code, draft and edit the manuscript,
and interpret experimental outputs. They were also the frozen systems under experimental
study. The authors reviewed the cited sources, protocol, code, analyses, and final artifacts and
take responsibility for their correctness.

\appendix
\section{Secondary selection and transfer diagnostics}
\label{app:secondary-diagnostics}

\begin{figure}[t]
    \centering
    \input{figures/fixed_trajectories.tex}
    \caption{Mean fixed-bank hidden utility on the eight fixed-order streams. Panel (a)
    gives the scope intervention and controls. Panel (b) compares permissive baselines
    against Static. Shading denotes stream-bootstrap 95\% intervals.}
    \label{fig:trajectory}
\end{figure}

\IfFileExists{generated_external.tex}{\input{generated_external.tex}}{}
\IfFileExists{generated_randomized_entry.tex}{\input{generated_randomized_entry.tex}}{}

We first ask whether the gate can identify updates that improve the current family without
regressing on previously seen families. Across 479 proposal--round comparisons, the public-test
component has AUROC 0.752 and accuracy 0.564 for this outcome. It marks 209 unsuccessful
comparisons as positive: 197 are harmful and 12 do not improve performance. The score based on
private and metamorphic tests raises AUROC to 0.998, leaving 12 unsuccessful positives (three
harmful and nine non-improving). Its errors have correlation $\phi=0.164$ with public-test errors.
At the diagnostic threshold, the score combining private and metamorphic tests gives the
same predictions when the public score is also required to pass, across these 479 fixed comparisons.
The separate online Scoped-Private ablation uses private test scores with program safety checks
and covers the 27 full-length streams in \cref{app:reviewer-controls}.

The local and global evaluations reveal different effects of the same skills. All eight
\method{} acceptances are safe on hidden tests of the same-family gate probes; two improve
the next global checkpoint and six reduce it. The descriptive positive predictive value therefore falls from
1.00 on local probes to 0.25 on the global checkpoint. At round 0, no historical family is
available to reveal these cross-family regressions to the gate.

For transfer beyond \bench, we evaluate final skills on 32 HumanEval+ tasks selected
before adaptive main results by a fixed SHA-256 ranking. Its augmented tests run offline
in a pinned NumPy container. A second diagnostic applies final skills to the original
18-task final banks with GPT-OSS-120B on AkashML BF16 and a 2,000-token cap. Both diagnostics
hold the skills fixed. On HumanEval+,
Global-\method{} scores 0.895 versus Static's 0.914. On the hidden GPT-OSS test banks,
Global-\method{} scores 0.951 and task-specific retrieval reaches 0.960. The scoped advantage is
0.020 over Static (bootstrap 95\% interval $[0.004,0.035]$) and 0.010 over Global-\method{}
($[-0.033,0.055]$). The latter interval includes zero.

\section{Proof of the gate-drift proposition}

Conditioning on whether a proposal is beneficial, the expected deployed change is
$\pi\alpha G-(1-\pi)\beta L$, positive exactly under the stated inequality. Under conditional
independence, all $m$ channels falsely accept a harmful proposal with probability
$\prod_j\beta_j$. If false-acceptance events are identical, their conjunction is the same event
and retains probability $\beta$.

\section{Evidence roles and leakage boundaries}

Each test set has a separate role. The editor receives summaries of discovery programs,
visible errors, specifications, and public examples. The actor solves probe tasks to compare
skills, but those tasks are never summarized for the editor. Only the gate receives private-test
and metamorphic outcomes.

Checkpoint scores are recorded after each decision and never enter proposal or acceptance
prompts. Hidden cases are absent from online manifests and loaded only by the offline evaluator,
which makes no model calls. The final test bank is separate from the checkpoint bank and is
used once to evaluate the final skill state.

\section{Examples of accepted skills}

In the eight main streams, every Global-\method{} run accepts only its round-0 boundary-family
proposal. The resulting skills
describe the boundary repair rule learned in the first round. For example, seed 101
ends with ``count values where \texttt{low <= x <= high}'' and seed 337 adds a rule to infer open
bounds when endpoints are excluded by examples. These rules follow the discovery
examples and pass hidden tests on the same-family gate probes. Their global checkpoint
scores show how the same rule affects other families when supplied to every task. The rule
then remains in use because the gate rejects later proposals. These examples illustrate the cross-family interference measured in the quantitative experiments.

\section{Pilot and protocol changes}

In the first paid pilot stream, Latest-only reached a perfect hidden checkpoint after one
update with the original v2 generator. These results were archived and excluded. The v3
diagnostic/neutral split restored initial hidden performance to the prespecified difficulty
range. Across two six-round v3 pilot streams, ORC accepted one of 12 proposals, so the default
threshold rejected most updates but still allowed an acceptance; we retained it. While generating task manifests for the main study, three declared seeds lacked eight neutral boundary inputs; before any
main call, v3.1 added explicit neutral and diagnostic constructors, updated the generator version,
and added a test that generates all eight complete streams. No task semantic, seed, gate, or
threshold changed.

The sequential editor input was fixed at 48,000 characters. Batch-\method~was initially given
the same limit, but an input-length check requiring no model calls found that accumulated feedback
for seed 137 required 52,487 characters. The completed seed-101 48,000-character batch run and a partial
seed-137 run were archived and excluded; the batch limit was raised uniformly to 64,000 and all
eight batch runs were restarted with that budget. The larger budget accommodates the complete feedback and is included in cost accounting.
The evidence and acceptance logic remain unchanged. It was recorded before
the remaining batch runs were executed. The complete dated log, including network-timeout
reruns and unique run identifiers, is distributed with the code.

\section{Full randomized-order extension details}

The extension uses 27 seeds selected without observing their outcomes. Exactly three streams
begin with each of the nine families. Eighteen seeds extend the one-update experiment and
reuse its corresponding early tasks and complete final bank. The remaining nine seeds
complete the first-family balance. The 27-stream follow-up compares methods on these
same streams. Within a seed, Global- and Scoped-\method{} use the same manifest,
endpoint, limits, and 12-round family order. An acceptance updates the single global skill for
Global-\method{} and only the current-family slot for Scoped-\method{}. The initial skill is the
fallback for a family with no accepted slot. We estimate uncertainty by resampling complete streams 50,000 times with a fixed seed.
Two-sided paired sign-flip tests use 100,000 Monte Carlo draws. Holm correction accounts for
testing the three utility contrasts together. The multiple-acceptance comparison uses the paired
stream indicator and is secondary.

In the original online task-specific-memory experiment, the task manifest supplies the family label used
to choose each skill slot.
In a retrospective checkpoint diagnostic, a simulated miss falls back to the initial skill.
At miss probability $q$, utility is the expectation $(1-q)s_{\mathrm{slot}}+q s_0$ over
archived oracle and initial-skill scores. Separately, a character 3--5-gram multinomial
Naive Bayes classifier trained on discovery and probe prompts from the other 26 seeds
classifies 243/243 held-out checkpoint prompts. That diagnostic retains the task-description
templates shared across seeds.

All 54 runs complete before hidden evaluation. Runs interrupted by network failures are
archived, charged to the shared ledger, and restarted from round zero with new run identifiers. No partial attempt enters the analysis. The extension's ledger-charged cost is
\$2.7527; total main, external, one-update, and full-extension spend is \$6.9965.

\section{Same-stream selection, capacity, and routing controls}
\label{app:reviewer-controls}

\IfFileExists{generated_reviewer_online.tex}{\input{generated_reviewer_online.tex}}{}

\paragraph{Static uses the same tasks and generation settings.}
Static uses the original skill throughout. We reuse its 243 checkpoint completions at all
13 checkpoint times, and its 486 distinct final-bank completions from the archived retrieval
tests. This gives a no-update reference on exactly the same tasks. The actor requests use the same task view, skill text, seed, provider,
temperature, and completion limit. Checkpoint and final-bank tasks are scored separately with their hidden tests. One duplicate checkpoint
dispatch produced different programs with identical hidden outcomes. Static consistently uses the
response stored for the exact request in the shared cache. The paired analyses use whole-stream bootstrap intervals with
50,000 draws. Holm adjustment covers the six
ORC-versus-Static tests and preserves the original three Scoped-versus-Global tests as a
separate family.

\begin{table}[ht]
\centering
\small
\setlength{\tabcolsep}{5pt}
\caption{Archived selection controls on 27 paired 12-round streams. Counts are harmful
acceptances among all acceptances and streams with at least one harmful acceptance.}
\label{tab:scoped-gates}
\begin{tabular}{lrrrrr}
\toprule
Gate & Trajectory & Final checkpoint & Final bank & Harmful & Streams \\
\midrule
Scoped-\method{} & 0.848 & 0.900 & 0.862 & 0/63 & 0/27 \\
Scoped-Private & 0.848 & 0.900 & 0.862 & 0/63 & 0/27 \\
Scoped-Latest & 0.871 & 0.945 & 0.940 & 24/321 & 20/27 \\
\bottomrule
\end{tabular}
\end{table}

\paragraph{Latest accepts more updates, including both gains and losses.}
Scoped-\method{} minus Scoped-Latest mean trajectory utility is
$-0.023$ ($[-0.043,-0.003]$, Holm $p=0.073$). Scoped-Private and Scoped-\method{} have
zero difference on all three utility measures. Private retains the original groupwise
private score rule and program-safety checks, while Latest accepts every well-formed
proposal that changes the current skill. The actor, editor, slots, seed manifests, and prompt limits are shared.
Scoped-Latest records 5,265 actor and editor requests, including cache hits, versus 22,113 each for
Scoped-Private and Scoped-\method{}. Additional ledger charges including retries are
\$0.231, \$0.005, and \$0.848, respectively, with differing cache reuse.

In all 258 rounds where decisions differ, Latest accepts and \method{} rejects. Of these,
26 rounds share the exact logged proposal, current skill, and pre-update memory state. All 26
fail the original gate's private gain check. Reapplying the private-only rule with the same
program-safety check to the exact archived \method{} gates changes zero decisions in each
324-round experiment, both task-specific and global. That offline diagnostic holds the proposals, probe
solutions, and histories fixed. The online Scoped-Private control reruns the full update process, allowing later proposals
to change. Five editor responses differ from Scoped-\method{}, but the acceptance decisions
and utility outcomes remain the same.

\paragraph{Replay accepts more proposals with the same family slots.}
Scoped-Replay compares pooled public means on the current probes and all previous probes.
Of 226 decisions that differ from Scoped-\method{}, 225 are Replay-only acceptances.
There are 26 differences on the exact same logged proposal, current skill, and prior memory,
all with a failed private gain check under \method{}. The sole \method{}-only acceptance
is the boundary decision at seed 1002, round index 9. The two runs propose different
skills there, and Replay's candidate lowers both public means. Replay gains 0.03585 in
trajectory utility ($[0.01990,0.05288]$, Holm $p=0.00014$) and 0.091 in final-bank utility
($[0.064,0.120]$).

\paragraph{Permissive gates perform differently with global memory.}
On the same 27 streams, Global-\method{} attains 0.785, compared with 0.696 for
Global-Latest and 0.744 for Global-Replay. The paired trajectory gains of
Global-\method{} are 0.089 ($[0.058,0.119]$, Holm $p=0.00006$) and 0.041
($[0.007,0.074]$, Holm $p=0.052$), respectively. Global-Latest and Global-Replay
each incur at least one harmful acceptance in all 27 streams. They differ from
Global-\method{} in 312 and 139 round-aligned decisions. The private-only gate shares
all 324 acceptance decisions with Global-\method{}. Its small utility difference
comes from a single stream with different model outputs. The descriptive scoped-minus-global
trajectory gain is 0.140 under Replay and 0.063 under \method{}.

\paragraph{Selecting one entry improves utility over presenting the whole shared bank.}
A separate 27-stream comparison gives both methods up to nine 360-character family skills and a
4,096-character padded actor memory component. Task-specific retrieval supplies one entry. The shared
prompt bank gives the actor all stored entries with family tags and an instruction to apply
only relevant rules. The paired trajectory difference is 0.056 ($[0.033,0.085]$, Holm
$p<10^{-4}$). That difference includes changes in proposals and acceptances. With final
skills held fixed and headers matched, presenting one tagged entry improves final-bank
utility by 0.028 ($[0.0001,0.059]$) over presenting all tagged entries. Both fixed-content
prompts supply the current family label. These comparisons match character limits and address
shared-prompt interference.

\paragraph{Top-1 retrieval uses the original active skill limit.}
The matched online comparison gives both stores nine entries of up to 360 characters,
3,240 stored learned-skill characters, a 360-character actor skill component, and 18 bytes
of metadata recording entry lengths. Both methods share the same read-only initial skill and editor instructions (meta-skill). The protocol fixes which slot belongs to each family. Family lookup replaces a skill in its family slot.
Multiversion retrieval chooses a skill by word overlap with the public task specification.
After removing direct protocol cues, it represents the presence of each term as a binary
feature and ranks skills by cosine similarity. It supplies the highest-ranked skill with positive similarity,
breaks ties by recency, and falls back to the initial skill if no match has positive score.
Adding an exact duplicate refreshes its recency; adding a new entry to a full store removes
the oldest entry. Before accepting an update, the same ORC gate tests the memory that would result from
adding the skill and, if needed, removing an old entry. This adapts the
workflow retrieval design in AgentMemoryBench \citep{ma2026benchmarking}, using local
lexical scoring and the same editor and evidence permissions as Scoped-\method{}.
The matched task-specific and multiversion arms retain a mean of 2.33 and 1.81 skills at
termination, with maxima of five and six, so both methods remain within the nine-entry cap.
The contrast's Holm-adjusted trajectory sensitivity test gives $p=0.00007$. Multiversion
incurs eight harmful acceptances across seven streams. Both methods record 22,113 calls, including cache hits. The matched task-specific arm reuses its cache. Multiversion costs \$0.227 including
the recovered attempts.

\paragraph{Wrong-skill retrieval differs from fallback to the initial skill.}
We apply every available wrong family skill to each archived final-bank task. For each task,
uniform wrong-slot utility averages populated nonmatching slots and uses the initial skill
when none is available. Wrong-slot minus initial-skill utility is $-0.077$
($[-0.114,-0.043]$). A second retrospective analysis uses the same leave-one-seed-out scheme
but replaces the task text with public argument and result shapes and sequence-length
relations. The router receives no names, code, template text, protocol tokens, or hidden
outcomes. It predicts the correct label on 409/486 final tasks, selects a populated correct
slot on 111 tasks and a populated wrong slot on 47, and falls back to the initial skill on
328. With final memory fixed, its mean final-bank utility is 0.818, a paired difference of
$-0.043$ ($[-0.059,-0.028]$) from the oracle. The public I/O patterns still reflect the
generated families. In the online control, the same frozen classifier selects read slots and
writes to the family predicted most often among confident discovery-task predictions,
provided that this family has a unique highest count. If retrieval ties or selects an empty slot, the actor uses the initial skill. If the write
vote ties, the editor makes no proposal. The gate still uses the known task-family labels to group its evaluation tasks. Later
proposals can change as the memory evolves.
Scoped-Learned attains a trajectory mean of 0.810 and final-bank mean of 0.805. It records
91 correct learned-slot reads, 43 wrong-slot reads, and 352 fallbacks on the 486 final tasks.
Across 3,159 checkpoint reads it records 173 wrong-slot reads and 2,637 fallbacks.
All 1,296 discovery labels are predicted correctly. The 46 accepted deployments
include eight immediate losses in eight streams.

\paragraph{The task-specific one-shot control delays every update until the final round.}
Scoped-Cumulative uses the original skill and slot budgets. It accumulates public discovery
feedback under the initial skill, then makes one proposal for each family after the last
discovery batch. The latest probe round of each family supplies its gain check, and the
remaining probes supply regression checks. Each proposed skill is tested independently against the initial skill. All accepted skills deploy together, so aggregate harm is counted once for the
combined deployment, while declines on a skill's own family are recorded separately for each insertion. Of the 243 family
proposals, 34 deploy across 15 streams. No harmful deployment is observed. The Holm
adjusted sensitivity tests for sequential minus cumulative utility give $p=0.00045$ at
the final checkpoint and $p=0.00028$ on the final bank. Cumulative uses 243 editor
calls versus 324 for Scoped-\method{}, and 28,512 total recorded calls versus 22,113.
Its gate evaluates more tasks per proposal. Summing the reported usage
for all recorded calls, including cache hits, gives \$3.753 versus \$3.088 at the original prices. Actual additional
charges are \$0.220 for this run, which reuses cached responses, and \$0.848 for the original sequential run.

\section{Full-pipeline second-backbone comparison}
\label{app:second-backbone}

GPT-OSS-120B serves as both actor and editor on the fixed Crusoe BF16 endpoint through
OpenRouter, using temperature zero, the seed recorded in the task manifest, and a 2,000-token actor allowance
and 4,096-token editor allowance inclusive of reasoning. Every method starts with the original skill
and the same editor instructions (meta-skill). The editor sees only public discovery feedback. The gate uses the original
public, private, and metamorphic checks. Task-specific retrieval uses the known family labels, and multiversion
retrieval uses the matched lexical workflow store in \cref{app:reviewer-controls}. Hidden evaluation starts only after all 108 online runs finish. The earlier AkashML actor diagnostic uses frozen final
skills (\cref{app:secondary-diagnostics}).

The paired contrasts use 50,000 whole-stream bootstrap samples and 100,000 sign-flip draws.
Holm correction is applied jointly to the three trajectory comparisons. Final-checkpoint and final-bank comparisons are descriptive.
\Cref{tab:second-backbone,tab:second-backbone-effects} give all estimates.
Static starts at 0.909 checkpoint utility, compared with 0.793 on Qwen. Across 972 editor
requests, including cache hits, 417 responses lack a parseable skill, including 387 that reach the generation cap.
Scoped and multiversion accept the same two normalization proposals and retain at most one
learned skill in any stream. Global-\method{} rejects one of those same proposals on its
historical boundary private nonregression check. This rerun characterizes adaptation from a high-performing initial skill, with sparse writes
and at most one populated slot per stream.

\input{generated_second_backbone.tex}

Together, the four methods record 48,272 calls, including cache hits. The provider charges for 8,200 completions for the four-arm experiment and 19 public test calls
used to check the endpoint. No charge has an unresolved billing status. The second-backbone supplement provides usage by method, paired stream results, and gate
decisions.

\section{Sandbox and budget checks}

The artifact tests sandbox isolation, timeouts, concurrent budget reservations, manifest
integrity, ORC logic, and backward transfer. A SQLite transaction reserves cost before
dispatch. The ledger records provider-reported charges. If a request times out after dispatch and its
billing status is unknown, the ledger charges the reserved amount conservatively. Successful raw responses are archived before parsing.

\end{document}

%% file: math_commands.tex
\usepackage{amsmath,amsfonts,bm}

\def\eqref#1{equation~\ref{#1}}
\def\1{\bm{1}}

\newcommand{\test}{\mathcal{D_{\mathrm{test}}}}

\DeclareMathAlphabet{\mathsfit}{\encodingdefault}{\sfdefault}{m}{sl}
\SetMathAlphabet{\mathsfit}{bold}{\encodingdefault}{\sfdefault}{bx}{n}

%% file: figures/figure_icons.tex
\newcommand{\paperfigureicon}[5][1]{%
    \begin{scope}[shift={(#3,#4)}, x=1mm, y=1mm, scale=#1, color=#5,
        line width=0.72pt, line cap=round, line join=round]
        \pic {paper icon #2};
    \end{scope}%
}

\tikzset{
    pics/paper icon user/.style={code={
        \draw (0,1.65) circle[radius=1.25mm];
        \draw (-2.95,-2.65) .. controls (-2.78,-1.02) and (-1.55,-0.38) .. (0,-0.38)
            .. controls (1.55,-0.38) and (2.78,-1.02) .. (2.95,-2.65);
    }},
    pics/paper icon agent/.style={code={
        \draw[rounded corners=0.8mm] (-2.75,-2.45) rectangle (2.75,1.75);
        \draw (0,1.75) -- (0,2.75);
        \draw (0,3.18) circle[radius=0.42mm];
        \draw (-3.35,0.3) -- (-3.35,-1.0) (3.35,0.3) -- (3.35,-1.0);
        \fill (-1.05,0.12) circle[radius=0.32mm];
        \fill (1.05,0.12) circle[radius=0.32mm];
        \draw (-0.9,-1.32) -- (0.9,-1.32);
    }},
    pics/paper icon edit/.style={code={
        \draw[rounded corners=0.3mm] (0,-3.05) -- (-2.95,-3.05) -- (-2.95,2.8)
            -- (1.0,2.8) -- (1.0,2.05);
        \draw (-1.9,1.35) -- (-0.72,1.35) (-1.9,0.12) -- (-1.05,0.12);
        \draw (-0.4,-2.2) -- (-0.03,-0.54) -- (2.42,1.9) -- (3.55,0.77)
            -- (1.1,-1.68) -- cycle;
        \draw (1.85,1.32) -- (2.98,0.19);
    }},
    pics/paper icon shield/.style={code={
        \draw (0,3.2) -- (2.85,2.05) -- (2.85,0.1)
            .. controls (2.85,-1.58) and (1.65,-2.62) .. (0,-3.18)
            .. controls (-1.65,-2.62) and (-2.85,-1.58) .. (-2.85,0.1)
            -- (-2.85,2.05) -- cycle;
        \draw (-1.28,0.12) -- (-0.32,-0.88) -- (1.45,1.03);
    }},
    pics/paper icon memory/.style={code={
        \foreach \level in {-2.65,-0.76,1.13} {
            \draw[rounded corners=0.35mm] (-3.0,\level) rectangle (3.0,{\level+1.46});
            \fill (-1.93,{\level+0.73}) circle[radius=0.20mm];
            \draw (-0.8,{\level+0.73}) -- (2.03,{\level+0.73});
        }
    }},
    pics/paper icon device/.style={code={
        \draw[rounded corners=0.75mm] (-2.22,-3.35) rectangle (2.22,3.35);
        \draw (-0.64,2.38) -- (0.64,2.38) (-0.52,-2.4) -- (0.52,-2.4);
        \draw (-0.72,0.68) -- (-1.38,0.04) -- (-0.72,-0.6)
              (0.72,0.68) -- (1.38,0.04) -- (0.72,-0.6);
    }},
    pics/paper icon discovery/.style={code={
        \draw[rounded corners=0.25mm] (-0.25,-3.0) -- (-3.0,-3.0) -- (-3.0,3.02)
            -- (1.28,3.02) -- (1.28,1.9);
        \draw (-2.05,1.78) -- (-0.2,1.78) (-2.05,0.65) -- (-1.03,0.65);
        \draw (1.05,-0.25) circle[radius=1.55mm];
        \draw (2.16,-1.35) -- (3.45,-2.7);
    }},
    pics/paper icon probes/.style={code={
        \draw[rounded corners=0.3mm] (-3.0,-1.4) -- (-3.0,3.1) -- (0.9,3.1) -- (0.9,1.42);
        \draw[rounded corners=0.3mm] (-0.97,-3.07) rectangle (3.0,1.42);
        \draw (-2.08,2.02) -- (-0.12,2.02)
              (-0.02,0.28) -- (2.05,0.28)
              (-0.02,-0.8) -- (2.05,-0.8)
              (-0.02,-1.88) -- (1.32,-1.88);
    }},
    pics/paper icon checkpoint/.style={code={
        \draw (-2.95,2.7) -- (-2.95,-2.65) -- (3.0,-2.65);
        \draw (-1.95,-1.25) -- (-0.64,0.22) -- (0.65,-0.26) -- (2.5,1.75);
        \foreach \spot in {(-1.95,-1.25),(-0.64,0.22),(0.65,-0.26),(2.5,1.75)}
            \fill \spot circle[radius=0.26mm];
    }},
    pics/paper icon flag/.style={code={
        \draw (-2,-3.05) -- (-2,3.05) -- (2.85,3.05) -- (1.65,1.47)
            -- (2.85,-0.02) -- (-2,-0.02);
        \draw (-3.05,-3.05) -- (-0.95,-3.05);
    }},
    pics/paper icon check/.style={code={
        \draw (0,0) circle[radius=2.65mm];
        \draw (-1.3,0) -- (-0.34,-0.97) -- (1.38,0.93);
    }},
    pics/paper icon keep/.style={code={
        \draw (-1.97,1.67) arc (140:-162:2.57mm);
        \draw (-2.68,2.32) -- (-2.68,0.85) -- (-1.2,0.85);
    }},
}

%% file: figures/trajectory_styles.tex
\definecolor{rsiScopeTeal}{HTML}{14665D}
\definecolor{rsiGlobalNavy}{HTML}{3A5575}
\definecolor{rsiStaticGray}{HTML}{414854}
\definecolor{rsiLatestOrange}{HTML}{B44C00}
\pgfplotsset{
  rsi trajectory axis/.style={
    scale only axis,
    width=.365\linewidth,
    height=1.43in,
    axis x line*=bottom,
    axis y line*=left,
    axis line style={rsiStaticGray,line width=.45pt},
    tick align=outside,
    tick style={black!65,line width=.4pt},
    major tick length=2pt,
    ymajorgrids,
    major grid style={rsiGlobalNavy!18,line width=.35pt},
    tick label style={font=\footnotesize},
    yticklabel style={
      /pgf/number format/fixed,
      /pgf/number format/fixed zerofill,
      /pgf/number format/precision=2,
    },
    label style={font=\footnotesize},
    title style={font=\small\bfseries,anchor=south west,inner sep=0pt},
    legend style={
      at={(0,1.025)},anchor=south west,draw=none,fill=none,
      font=\footnotesize,cells={anchor=west},inner sep=0pt,
      /tikz/column 2/.style={column sep=2pt},
    },
  },
  rsi scoped line/.style={
    rsiScopeTeal,line width=.95pt,mark=*,mark size=1.55pt
  },
  rsi global line/.style={
    rsiGlobalNavy,line width=.9pt,dashed,mark=square*,mark size=1.35pt
  },
  rsi static line/.style={
    rsiStaticGray,line width=.9pt,densely dotted,mark=none
  },
  rsi latest line/.style={
    rsiLatestOrange,line width=.9pt,dashed,mark=square*,mark size=1.35pt
  },
  rsi replay line/.style={
    rsiGlobalNavy,line width=.9pt,mark=triangle*,mark size=1.9pt
  },
}
\newcommand{\rsitrajectoryband}[4]{%
  \addplot[name path=rsiLower,draw=none,forget plot]
    table[x=#2,y=#3Low] #1;
  \addplot[name path=rsiUpper,draw=none,forget plot]
    table[x=#2,y=#3High] #1;
  \addplot[draw=none,fill=#4,fill opacity=.12,forget plot]
    fill between[of=rsiLower and rsiUpper];
}

%% file: figures/agent_memory_mechanism.tex
\definecolor{memoryInk}{HTML}{202E3A}
\definecolor{memoryBorder}{HTML}{A6B5BE}
\definecolor{memorySurface}{HTML}{F4F7F9}
\definecolor{memoryTeal}{HTML}{14665D}
\definecolor{memoryHighlight}{HTML}{E5F3EF}
\definecolor{memoryNavy}{HTML}{3A5575}
\definecolor{memoryBadge}{HTML}{E6EDF5}

\begin{tikzpicture}[
    x=1mm, y=1mm,
    every node/.style={inner sep=0pt, outer sep=0pt, text=memoryInk, font=\footnotesize},
    heading/.style={anchor=west, font=\small\bfseries},
    card/.style={rounded corners=2.5pt, draw=memoryBorder, fill=memorySurface},
    accepted/.style={card, draw=memoryTeal, fill=memoryHighlight},
    badge/.style={fill=memoryBadge},
    connector/.style={->, >=latex, draw=memoryInk, line width=0.65pt},
    title/.style={anchor=west, font=\footnotesize\bfseries},
    copy/.style={anchor=north west, align=left},
]
    \node[heading] at (0,0) {1. Update memory on family $f$};

    \draw[card] (0,-4) rectangle (38,-22);
    \fill[badge] (5.85,-8.7) circle[radius=3.45mm];
    \paperfigureicon[0.78]{agent}{5.85}{-8.7}{memoryNavy}
    \node[title] at (11,-8.7) {Frozen actor};
    \node[copy] at (2.5,-13.8) {Repair discovery tasks\\with active skill $K_t(f)$.};
    \draw[connector] (38,-13) -- (45,-13);

    \draw[card] (45,-4) rectangle (83,-22);
    \fill[badge] (50.85,-8.7) circle[radius=3.45mm];
    \paperfigureicon[0.78]{edit}{50.85}{-8.7}{memoryNavy}
    \node[title] at (56,-8.7) {Frozen editor};
    \node[copy] at (47.5,-13.8) {Use visible feedback to\\propose a skill revision.};
    \draw[connector] (83,-13) -- (90,-13);

    \draw[card] (90,-4) rectangle (132,-22);
    \fill[badge] (95.85,-8.7) circle[radius=3.45mm];
    \paperfigureicon[0.78]{shield}{95.85}{-8.7}{memoryNavy}
    \node[title] at (101,-8.7) {ORC gate};
    \node[copy] at (92.5,-13.8) {Compare both skills on\\current and historical probes.};

    \draw[connector] (111,-22) -- (111,-25) -- (32,-25) -- (32,-28);
    \draw[connector] (111,-25) -- (111,-28);

    \draw[accepted] (0,-28) rectangle (64,-36);
    \paperfigureicon[0.78]{check}{5.4}{-32}{memoryTeal}
    \node[anchor=west] at (10,-32) {\textbf{Accept} \enspace Write to slot $f$ only.};
    \draw[card] (68,-28) rectangle (132,-36);
    \paperfigureicon[0.78]{keep}{73.4}{-32}{memoryNavy}
    \node[anchor=west] at (78,-32) {\textbf{Reject} \enspace Keep the stored skills.};

    \node[heading] at (0,-42) {2. Retrieve a skill for a later task};
    \draw[connector, draw=memoryTeal] (56,-36) -- (56,-46);

    \draw[card] (0,-46) rectangle (34,-69);
    \fill[badge] (5.85,-50.8) circle[radius=3.45mm];
    \paperfigureicon[0.78]{user}{5.85}{-50.8}{memoryNavy}
    \node[title] at (11,-50.8) {Incoming task};
    \node[copy] at (2.5,-56.7) {Boundary repair task\\Family ID $f$};
    \draw[connector] (34,-57.6) -- (41,-57.6);

    \draw[card] (41,-46) rectangle (91,-69);
    \fill[memoryHighlight] (46.85,-50.8) circle[radius=3.45mm];
    \paperfigureicon[0.78]{memory}{46.85}{-50.8}{memoryTeal}
    \node[title] at (52,-50.8) {Family memory};
    \draw[accepted] (43.5,-55.1) rectangle (88.5,-60.6);
    \node[anchor=west] at (45.5,-57.85) {Boundary ($f$)};
    \node[anchor=east, text=memoryTeal] at (86.5,-57.85) {\textbf{Matched}};
    \draw[card,fill=white] (43.5,-61.6) rectangle (88.5,-67.1);
    \node[anchor=west] at (45.5,-64.35) {Ordering ($g$)};
    \node[anchor=east] at (86.5,-64.35) {Stored};
    \draw[connector, draw=memoryTeal] (91,-57.85) -- (98,-57.85);

    \draw[card] (98,-46) rectangle (132,-69);
    \fill[badge] (103.85,-50.8) circle[radius=3.45mm];
    \paperfigureicon[0.78]{device}{103.85}{-50.8}{memoryNavy}
    \node[title] at (109,-50.8) {Actor prompt};
    \node[copy] at (100.5,-56.7) {Task + matching skill\\Empty slot: use $K_0$.};
\end{tikzpicture}

%% file: figures/experiment_protocol.tex
\definecolor{protocolInk}{HTML}{202E3A}%
\definecolor{protocolBorder}{HTML}{A6B5BE}%
\definecolor{protocolSurface}{HTML}{F4F7F9}%
\definecolor{protocolTeal}{HTML}{14665D}%
\definecolor{protocolHighlight}{HTML}{E5F3EF}%
\begin{tikzpicture}[
    x=1mm, y=1mm,
    every node/.style={inner sep=0pt, outer sep=0pt, font=\footnotesize, text=protocolInk},
    heading/.style={anchor=west, font=\small\bfseries},
    task card/.style={rounded corners=2pt, draw=protocolBorder, fill=protocolSurface},
    evaluation card/.style={rounded corners=2pt, draw=protocolTeal, fill=protocolHighlight},
    flow/.style={->, >=latex, draw=protocolInk, line width=0.6pt},
    evaluation/.style={->, >=latex, draw=protocolTeal, densely dashed, line width=0.6pt},
    card title/.style={anchor=west, font=\footnotesize\bfseries},
    card text/.style={anchor=north west, align=left},
    icon badge/.style={fill=protocolHighlight},
]
    \node[heading] at (0,0) {Gated update within one task family};

    \draw[task card] (0,-4.5) rectangle (39,-22.5);
    \fill[icon badge] (7.4,-9.6) circle[radius=3.7mm];
    \paperfigureicon[0.82]{discovery}{7.4}{-9.6}{protocolTeal}
    \node[card title] at (13.5,-9.6) {Discovery};
    \node[card text] at (4,-14.2) {
        4 tasks supply visible\\feedback for the proposal};
    \draw[flow] (39,-13.5) -- (45,-13.5);

    \draw[task card] (45,-4.5) rectangle (93,-22.5);
    \fill[icon badge] (52.4,-9.6) circle[radius=3.7mm];
    \paperfigureicon[0.82]{probes}{52.4}{-9.6}{protocolTeal}
    \node[card title] at (58.5,-9.6) {Paired probes};
    \node[card text] at (49,-14.2) {
        4 new tasks and historical replay\\test current and proposed skills};
    \draw[flow] (93,-13.5) -- (99,-13.5);

    \draw[task card] (99,-4.5) rectangle (136,-22.5);
    \fill[icon badge] (106.4,-9.6) circle[radius=3.7mm];
    \paperfigureicon[0.82]{shield}{106.4}{-9.6}{protocolTeal}
    \node[card title] at (112.5,-9.6) {Update decision};
    \node[card text] at (103,-14.2) {
        Accept proposal or\\keep prior memory};

    \node[heading] at (0,-28.5) {Across 12 rounds with recurring families};

    % The dashed paths score each deployed state using separate evaluation tasks.
    \draw[evaluation card] (0,-46) rectangle (105,-65.5);
    \draw[evaluation card] (108.5,-46) rectangle (136,-65.5);

    \draw[flow] (14,-39.5) -- (38,-39.5);
    \draw[flow] (38,-39.5) -- (61,-39.5);
    \draw[flow] (86,-39.5) -- (98,-39.5);
    \draw[protocolInk, line width=0.6pt] (61,-39.5) -- (70,-39.5);
    \node at (78,-39.5) {$\cdots$};

    \foreach \x/\state/\score in {14/{Initial state}/0,38/{Round 1}/1,61/{Round 2}/2,98/{Round 12}/12} {
        \node at (\x,-35) {\state};
        \draw[evaluation] (\x,-39.5) -- (\x,-46);
        \fill[protocolInk] (\x,-39.5) circle[radius=0.65mm];
        \node[text=protocolTeal] at (\x,-50) {$U_{\score}$};
    }
    \draw[evaluation] (98,-39.5) -- (122.25,-39.5) -- (122.25,-46);
    \node[text=protocolTeal] at (78,-50) {$\cdots$};

    \fill[white] (30,-58.5) circle[radius=3.7mm];
    \paperfigureicon[0.82]{checkpoint}{30}{-58.5}{protocolTeal}
    \node[card title, text=protocolTeal] at (36.2,-56.5) {Fixed hidden checkpoint bank};
    \node[anchor=west] at (36.2,-60.5) {Same 9 tasks scored at every state};

    \fill[white] (113.85,-51.5) circle[radius=3.7mm];
    \paperfigureicon[0.82]{flag}{113.85}{-51.5}{protocolTeal}
    \node[card title, text=protocolTeal] at (119.35,-51.5) {Final bank};
    \node[align=center] at (122.25,-60.2) {18 tasks\\Scored once};
\end{tikzpicture}%

%% file: generated_results.tex
\begin{table}[t]
\centering
\small
\setlength{\tabcolsep}{3pt}
\caption{Main ProcStream-RSI results over eight paired streams. Brackets are seed-bootstrap 95\% intervals. Harmful accepts count accepted rounds whose fixed hidden checkpoint score decreased.}
\label{tab:main-results}
\resizebox{\linewidth}{!}{%
\begin{tabular}{lccccc}
\toprule
Method & Mean traj. $\uparrow$ & Final checkpoint $\uparrow$ & BWT $\uparrow$ & Accepted & Harmful \\
\midrule
Static & 0.775 [0.734, 0.814] & 0.775 [0.734, 0.814] & 0.000 [0.000, 0.000] & 0.00 & 0.00 \\
Frozen-compute & 0.775 [0.734, 0.814] & 0.775 [0.734, 0.814] & 0.000 [0.000, 0.000] & 0.00 & 0.00 \\
Latest-only & 0.703 [0.678, 0.725] & 0.720 [0.681, 0.755] & -0.253 [-0.300, -0.204] & 12.00 & 6.25 \\
Self-judge & 0.736 [0.696, 0.774] & 0.722 [0.641, 0.802] & -0.140 [-0.220, -0.059] & 3.38 & 1.75 \\
Replay & 0.707 [0.676, 0.738] & 0.681 [0.621, 0.745] & -0.240 [-0.297, -0.191] & 7.88 & 3.88 \\
ORC & 0.713 [0.653, 0.761] & 0.707 [0.645, 0.760] & 0.000 [0.000, 0.000] & 1.00 & 0.75 \\
Batch-ORC & 0.775 [0.734, 0.814] & 0.775 [0.734, 0.814] & 0.000 [0.000, 0.000] & 0.00 & 0.00 \\
\bottomrule
\end{tabular}%
}
\end{table}

\begin{table}[t]
\centering
\small
\caption{Prespecified paired contrasts. Intervals are descriptive percentile seed-bootstrap 95\% intervals. Paired sign-flip sensitivity $p$-values require symmetry of seed-level differences; Holm adjustment covers the three rows.}
\label{tab:contrasts}
\begin{tabular}{lccc}
\toprule
Contrast & Mean difference [bootstrap 95\%] & $p$ & Holm $p$ \\
\midrule
ORC - Replay mean trajectory & 0.006 [-0.045, 0.048] & 0.8438 & 1.0000 \\
ORC - Latest-only mean trajectory & 0.010 [-0.036, 0.051] & 0.7188 & 1.0000 \\
ORC - Batch-ORC final & -0.068 [-0.123, -0.004] & 0.0938 & 0.2812 \\
\bottomrule
\end{tabular}
\end{table}

%% file: generated_scoped.tex
\begin{table}[t]
\centering
\small
\caption{Same accepted skills, different retrieval scope on the eight main streams. Task-specific retrieval applies each ORC rule only to its originating family and uses the initial skill elsewhere.}
\label{tab:scoped-retrieval}
\begin{tabular}{lcccc}
\toprule
Deployment & Mean traj. $\uparrow$ & Final checkpoint $\uparrow$ & Accepted & Harmful \\
\midrule
Static & 0.775 [0.734, 0.814] & 0.775 [0.734, 0.814] & 0 & 0 \\
Global ORC & 0.713 [0.653, 0.762] & 0.707 [0.645, 0.760] & 8 & 6 \\
Task-specific retrieval & 0.816 [0.782, 0.852] & 0.819 [0.786, 0.855] & 8 & 0 \\
\bottomrule
\end{tabular}
\end{table}

%% file: figures/family_interference.tex
\definecolor{interferenceInk}{HTML}{202E3A}
\definecolor{interferenceGlobal}{HTML}{3A5575}
\definecolor{interferenceScoped}{HTML}{14665D}
\definecolor{interferenceGrid}{HTML}{DCE4E8}
\definecolor{interferenceHighlight}{HTML}{E5F3EF}

\begin{tikzpicture}[
    x=1mm, y=1mm,
    every node/.style={inner sep=0pt, outer sep=0pt, text=interferenceInk, font=\footnotesize},
    global line/.style={draw=interferenceGlobal, line width=0.85pt},
    scoped line/.style={draw=interferenceScoped, line width=0.85pt, densely dashed},
]
    \path[use as bounding box] (0,-60) rectangle (136,0);
    \draw[global line] (28,-3) -- (36,-3);
    \fill[interferenceGlobal] (32,-3) circle[radius=1.8pt];
    \node[anchor=west] at (38,-3) {Global retrieval};
    \draw[scoped line] (77,-3) -- (85,-3);
    \filldraw[fill=white, draw=interferenceScoped] (81,-3) circle[radius=1.8pt];
    \node[anchor=west] at (87,-3) {Task-specific retrieval};

    \fill[interferenceHighlight, rounded corners=1.5pt] (0,-7.5) rectangle (135,-12.5);
    \foreach \tick/\tickLabel in {30/-0.5, 50/-0.3, 70/-0.1, 90/0.1, 110/0.3, 130/0.5} {
        \draw[interferenceGrid] (\tick,-7.5) -- (\tick,-53.5);
        \node[anchor=north] at (\tick,-54) {$\tickLabel$};
    }
    \draw[interferenceInk!65] (80,-7.5) -- (80,-53.5);
    \node[anchor=north] at (80,-54) {$0$};

    % Global deltas are from results/main_summary.json (orc_round0_mean_delta_by_family).
    % The scoped deltas apply the archived-completion splice in analyze_scoped_counterfactual.py.
    \foreach \family/\height/\globalDelta/\scopedDelta in {
        \textbf{Boundary}/-10/0.3984375/0.3984375,
        Nested/-15/0.1171875/0,
        Ordering/-20/0/0,
        Normalization/-25/-0.046875/0,
        Chunk/-30/-0.1171875/0,
        Prefix/-35/-0.1171875/0,
        Tie-breaking/-40/-0.1484375/0,
        Rotation/-45/-0.234375/0,
        Missing/-50/-0.4609375/0
    } {
        \pgfmathsetmacro{\globalPoint}{80+100*\globalDelta}
        \pgfmathsetmacro{\scopedPoint}{80+100*\scopedDelta}
        \node[anchor=west] at (2,\height) {\family};
        \draw[global line] (80,{\height+1}) -- (\globalPoint,{\height+1});
        \fill[interferenceGlobal] (\globalPoint,{\height+1}) circle[radius=1.8pt];
        \draw[scoped line] (80,{\height-1}) -- (\scopedPoint,{\height-1});
        \filldraw[fill=white, draw=interferenceScoped]
            (\scopedPoint,{\height-1}) circle[radius=1.8pt];
    }
    \node[anchor=north] at (80,-58.5) {Mean change in hidden checkpoint utility};
\end{tikzpicture}

%% file: generated_randomized_entry_prose.tex
\paragraph{Varying the first task family.}
The one-update experiment tests all nine possible first families, with two streams per family.
The shared gate accepts 5 candidates across the 18 streams. Global use causes 3 harmful updates,
whereas task-specific use causes none. Mean hidden checkpoint utility after the decision is 0.761 for
Global-\method{} and 0.802 for Scoped-\method{}. The paired difference is 0.041
[-0.004, 0.107] (sign-flip sensitivity $p=0.1875$). The checkpoint interval includes both a small negative effect and a positive gain. The final-bank difference is 0.031 [0.001, 0.070]
($p=0.1250$).

The two methods share the accepted candidate and therefore have the same change on the current
family. Their mean changes on the other families are -0.047 for Global and 0.000 for Scoped.
This comparison isolates the effect of using a skill outside its originating family: scoped
retrieval preserves the current-family result without changing outputs on the other families.

%% file: generated_static_comparison.tex
\begin{table}[t]
\centering
\small
\setlength{\tabcolsep}{4.5pt}
\caption{Full 12-round randomized-order extension on 27 paired streams. Brackets give
stream-bootstrap 95\% intervals. Harm counts include accepted updates that reduce the next
hidden checkpoint and streams with at least one such update.}
\label{tab:randomized-full}
\resizebox{\linewidth}{!}{%
\begin{tabular}{lccccc}
\toprule
Method & Trajectory & Final checkpoint & Final bank & Harmful/accepted & Harm streams \\
\midrule
Static & 0.793 [0.768, 0.818] & 0.793 [0.768, 0.818] & 0.764 [0.741, 0.788] & -- & 0/27 \\
Global-\method{} & 0.785 [0.757, 0.812] & 0.789 [0.758, 0.817] & 0.743 [0.710, 0.773] & 6/12 & 6/27 \\
Scoped-\method{} & 0.848 [0.825, 0.871] & 0.900 [0.871, 0.927] & 0.862 [0.830, 0.893] & 0/63 & 0/27 \\
\bottomrule
\end{tabular}}
\end{table}

%% file: figures/randomized_trajectories.tex
% Series from results/randomized_full_trajectories.json and results/router_error_curve.csv.

\begingroup
\pgfplotstableread{
round globalMean globalLow globalHigh scopedMean scopedLow scopedHigh
0 0.79320987654321 0.7674897119341564 0.8181584362139918 0.79320987654321 0.7674897119341564 0.8179012345679012
1 0.7680041152263374 0.7294238683127572 0.8037551440329218 0.8076131687242799 0.7824074074074074 0.8323045267489712
2 0.7816358024691359 0.7471707818930041 0.8132716049382716 0.8125 0.7872942386831275 0.8369341563786008
3 0.7770061728395062 0.7435699588477367 0.8083847736625515 0.8186728395061729 0.7934670781893004 0.8431069958847737
4 0.7821502057613169 0.7497427983539096 0.8117283950617284 0.8343621399176955 0.8076131687242799 0.8598251028806585
5 0.7821502057613169 0.7497427983539096 0.8122427983539096 0.841820987654321 0.8176440329218108 0.8649691358024691
6 0.7890946502057614 0.7582304526748971 0.8171296296296297 0.8513374485596708 0.8263888888888888 0.8762860082304527
7 0.7890946502057614 0.7582304526748971 0.8166152263374485 0.8539094650205762 0.8284465020576133 0.8788580246913581
8 0.7890946502057614 0.7579732510288066 0.8168724279835392 0.8647119341563786 0.8387345679012346 0.8896604938271605
9 0.7890946502057614 0.7582304526748971 0.8166152263374485 0.8711419753086419 0.8459362139917695 0.8948045267489712
10 0.7890946502057614 0.7584876543209877 0.8168724279835392 0.8845164609053497 0.856224279835391 0.911522633744856
11 0.7890946502057614 0.7579732510288066 0.8171296296296297 0.8950617283950618 0.8659979423868313 0.9223251028806585
12 0.7890946502057614 0.7582304526748971 0.8168724279835392 0.8999485596707819 0.8706275720164608 0.9272119341563786
}\rsiRandomized

\pgfplotstableread{
miss fallbackMean fallbackLow fallbackHigh
0.0 0.8483697372586262 0.8244697689142134 0.870588002532447
0.05 0.8456117442228553 0.8219234726179171 0.8680060937005382
0.1 0.8428537511870845 0.819369262424818 0.8648840614118392
0.15 0.8400957581513138 0.8168358262108263 0.8623189696106364
0.2 0.8373377651155429 0.8142766698322255 0.859267173156062
0.25 0.8345797720797721 0.8112337765115543 0.8565507280785059
0.3 0.8318217790440013 0.8085786641342196 0.8538382399493512
0.35 0.8290637860082305 0.8056455761316873 0.8510416666666666
0.4 0.8263057929724597 0.8033119658119658 0.8484647040202595
0.45 0.8235477999366888 0.8000959559987337 0.8460806426084204
0.5 0.8207898069009181 0.7974537037037037 0.843215811965812
0.55 0.8180318138651472 0.7944582937638494 0.8404410018993352
0.6 0.8152738208293764 0.7916824944602723 0.8377888572333018
0.65 0.8125158277936056 0.7884922048116493 0.8353553339664451
0.7 0.8097578347578348 0.7858875435264324 0.832710113960114
0.75 0.806999841722064 0.7829861111111112 0.8302469135802469
0.8 0.8042418486862931 0.7797325102880658 0.8277619499841722
0.85 0.8014838556505224 0.7767222617917062 0.8253442545109212
0.9 0.7987258626147515 0.7735675846786957 0.8227484963596075
0.95 0.7959678695789807 0.7706008626147515 0.8202823282684394
1.0 0.79320987654321 0.7672325102880658 0.8181584362139918
}\rsiRouter

\begin{tikzpicture}
\begin{groupplot}[
  rsi trajectory axis,
  group style={group size=2 by 1,horizontal sep=.135\linewidth},
  title style={at={(axis description cs:0,1.25)}},
]
\nextgroupplot[
  title={(a) Deployed trajectories},
  xlabel={Deployment round},
  ylabel={Hidden checkpoint score},
  xmin=-.22,xmax=12.22,xtick={0,4,8,12},
  ymin=.70,ymax=.95,ytick={.70,.75,.80,.85,.90,.95},
]
\rsitrajectoryband{\rsiRandomized}{round}{global}{rsiGlobalNavy}
\rsitrajectoryband{\rsiRandomized}{round}{scoped}{rsiScopeTeal}
\addplot[rsi scoped line] table[x=round,y=scopedMean] \rsiRandomized;
\addlegendentry{Scoped-ORC}
\addplot[rsi global line] table[x=round,y=globalMean] \rsiRandomized;
\addlegendentry{Global-ORC}

\nextgroupplot[
  title={(b) Routing sensitivity},
  xlabel={Simulated routing miss rate},
  ylabel={Mean trajectory utility},
  ylabel style={at={(axis description cs:-.15,.5)},anchor=south},
  xmin=-.02,xmax=1.02,xtick={0,.25,.5,.75,1},
  xticklabels={0\%,25\%,50\%,75\%,100\%},
  ymin=.70,ymax=.90,ytick={.70,.75,.80,.85,.90},
]
\rsitrajectoryband{\rsiRouter}{miss}{fallback}{rsiGlobalNavy}
\addplot[rsiGlobalNavy,line width=.95pt]
  table[x=miss,y=fallbackMean] \rsiRouter;
\addlegendentry{Simulated fallback}
\addplot[rsiScopeTeal,only marks,mark=diamond*,mark size=2.25pt]
  coordinates {(0,0.8483697372586262)};
\addlegendentry{Oracle = retrospective router}
\end{groupplot}
\end{tikzpicture}
\endgroup

%% file: figures/memory_controls.tex
\begingroup
\definecolor{controlInk}{HTML}{202E3A}
\definecolor{controlGrid}{HTML}{E1E7EB}
\definecolor{controlAxis}{HTML}{788C98}
\definecolor{controlTeal}{HTML}{14665D}

\begin{tikzpicture}[
    x=1mm, y=1mm,
    every node/.style={inner sep=0pt, outer sep=0pt, text=controlInk, font=\footnotesize},
    header/.style={font=\footnotesize\bfseries},
    interval/.style={draw=controlTeal, line width=0.8pt},
]
    \node[header, anchor=west] at (0,0) {Comparison and outcome};
    \node[header] at (84.5,0) {Difference in utility};
    \node[header, anchor=east] at (136,0) {Estimate};
    \draw[controlAxis] (0,-3.6) -- (136,-3.6);

    \begin{scope}[shift={(53,0)}, x=630mm]
        \foreach \tick in {0.02,0.04,0.06,0.08,0.10}
            \draw[controlGrid] (\tick,-5.2) -- (\tick,-68.5);
        \draw[controlAxis, densely dashed] (0,-5.2) -- (0,-68.5);
        \draw[controlAxis] (0,-68.5) -- (0.1,-68.5);
        \foreach \tick in {0,0.02,0.04,0.06,0.08,0.10} {
            \draw[controlAxis] (\tick,-68.5) -- (\tick,-69.8);
            \node at (\tick,-73) {$\tick$};
        }
    \end{scope}

    \node[header, anchor=west] at (0,-8) {Family slots vs. lexical top-1};
    \draw[controlGrid] (0,-27.5) -- (136,-27.5);
    \node[header, anchor=west] at (0,-30) {Known labels vs. learned routing};
    \draw[controlGrid] (0,-49.5) -- (136,-49.5);
    \node[header, anchor=west] at (0,-52) {Sequential vs. one-shot writing};

    % Paired means and bootstrap bounds, respectively, from the paired_utility
    % entries in results/reviewer_{memory,routing,cumulative}_summary.json.
    % For cumulative writing, show the two prespecified final outcomes.
    \foreach \metric/\row/\estimate/\lower/\upper in {
        Trajectory/-14/0.026/0.012/0.043,
        Final checkpoint/-19.5/0.051/0.026/0.079,
        Final bank/-25/0.060/0.026/0.099,
        Trajectory/-36/0.039/0.026/0.052,
        Final checkpoint/-41.5/0.067/0.048/0.086,
        Final bank/-47/0.057/0.040/0.075,
        Final checkpoint/-58/0.051/0.026/0.078,
        Final bank/-63.5/0.041/0.023/0.062
    } {
        \node[anchor=west] at (4,\row) {\metric};
        \node[anchor=east] at (136,\row) {$+\estimate$};
        \begin{scope}[shift={(53,0)}, x=630mm]
            \draw[interval] (\lower,\row) -- (\upper,\row);
            \draw[interval] (\lower,\row) ++(0,-0.9) -- ++(0,1.8);
            \draw[interval] (\upper,\row) ++(0,-0.9) -- ++(0,1.8);
            \fill[controlTeal] (\estimate,\row) circle[radius=2pt];
        \end{scope}
    }
\end{tikzpicture}
\endgroup

%% file: figures/fixed_trajectories.tex
% Series and bootstrap intervals from scripts/render_trajectory.py (trajectories).

\begingroup
\pgfplotstableread{
round staticMean staticLow staticHigh globalMean globalLow globalHigh scopedMean scopedLow scopedHigh latestMean latestLow latestHigh replayMean replayLow replayHigh
0 0.7751736111111112 0.734375 0.8142361111111112 0.7751736111111112 0.734375 0.8142361111111112 0.7751736111111112 0.734375 0.8142361111111112 0.7751736111111112 0.734375 0.8142361111111112 0.7751736111111112 0.734375 0.8142361111111112
1 0.7751736111111112 0.734375 0.8142361111111112 0.7074652777777778 0.6449652777777778 0.7604166666666666 0.8194444444444444 0.7855902777777778 0.8550347222222222 0.7074652777777778 0.6449652777777778 0.7604166666666666 0.7074652777777778 0.6449652777777778 0.7604166666666666
2 0.7751736111111112 0.7335069444444444 0.8142361111111112 0.7074652777777778 0.6449652777777778 0.7604166666666666 0.8194444444444444 0.7855902777777778 0.8550347222222222 0.7378472222222222 0.6788194444444444 0.8046875 0.7352430555555556 0.6744791666666666 0.8038194444444444
3 0.7751736111111112 0.734375 0.8142361111111112 0.7074652777777778 0.6440972222222222 0.7604166666666666 0.8194444444444444 0.7855902777777778 0.8550347222222222 0.671875 0.6215277777777778 0.7222222222222222 0.7100694444444444 0.6467013888888888 0.779513888888889
4 0.7751736111111112 0.7335069444444444 0.8142361111111112 0.7074652777777778 0.6449652777777778 0.7604166666666666 0.8194444444444444 0.7855902777777778 0.8550347222222222 0.7387152777777778 0.670138888888889 0.8064236111111112 0.6788194444444444 0.5850694444444444 0.7630208333333334
5 0.7751736111111112 0.734375 0.8142361111111112 0.7074652777777778 0.6440972222222222 0.7604166666666666 0.8194444444444444 0.7855902777777778 0.8559027777777778 0.5555555555555556 0.5112847222222222 0.6067708333333333 0.6553819444444444 0.5503472222222222 0.7534722222222222
6 0.7751736111111112 0.7335069444444444 0.8142361111111112 0.7074652777777778 0.6449652777777778 0.7604166666666666 0.8194444444444444 0.7855902777777778 0.8550347222222222 0.6640625 0.5807291666666666 0.7395833333333334 0.6597222222222222 0.5590277777777778 0.7447916666666666
7 0.7751736111111112 0.734375 0.8142361111111112 0.7074652777777778 0.6458333333333334 0.7612847222222222 0.8194444444444444 0.7855902777777778 0.8550347222222222 0.7335069444444444 0.657986111111111 0.8046875 0.7178819444444444 0.6770833333333333 0.7612847222222222
8 0.7751736111111112 0.734375 0.8142361111111112 0.7074652777777778 0.6440972222222222 0.7612847222222222 0.8194444444444444 0.7864583333333333 0.8550347222222222 0.7942708333333334 0.7439236111111112 0.8394097222222222 0.7526041666666666 0.6909722222222222 0.8098958333333333
9 0.7751736111111112 0.732638888888889 0.8142361111111112 0.7074652777777778 0.6449652777777778 0.7604166666666666 0.8194444444444444 0.7855902777777778 0.8550347222222222 0.704861111111111 0.6232638888888888 0.7777777777777778 0.7369791666666666 0.6649305555555556 0.8046875
10 0.7751736111111112 0.7335069444444444 0.8142361111111112 0.7074652777777778 0.6449652777777778 0.7604166666666666 0.8194444444444444 0.7864583333333334 0.8550347222222222 0.6822916666666666 0.6302083333333333 0.7352430555555556 0.6927083333333334 0.6354166666666667 0.7482638888888888
11 0.7751736111111112 0.7335069444444444 0.8151041666666666 0.7074652777777778 0.6449652777777778 0.7604166666666666 0.8194444444444444 0.7855902777777778 0.8550347222222222 0.6484375 0.5885416666666666 0.7083333333333334 0.6901041666666666 0.623263888888889 0.7560763888888888
12 0.7751736111111112 0.7335069444444444 0.8142361111111112 0.7074652777777778 0.6440972222222222 0.7612847222222222 0.8194444444444444 0.7855902777777778 0.8550347222222222 0.7196180555555556 0.6805555555555556 0.7560763888888888 0.6805555555555556 0.6206597222222222 0.7447916666666667
}\rsiFixed

\begin{tikzpicture}
\begin{groupplot}[
  rsi trajectory axis,
  group style={group size=2 by 1,horizontal sep=.135\linewidth},
  title style={at={(axis description cs:0,1.35)}},
  xmin=-.22,xmax=12.22,xtick={0,4,8,12},
  ymin=.50,ymax=.90,ytick={.50,.60,.70,.80,.90},
  xlabel={Deployment round},
]
\nextgroupplot[
  title={(a) Scope and controls},
  ylabel={Fixed-bank hidden utility},
]
\rsitrajectoryband{\rsiFixed}{round}{static}{rsiStaticGray}
\rsitrajectoryband{\rsiFixed}{round}{global}{rsiGlobalNavy}
\rsitrajectoryband{\rsiFixed}{round}{scoped}{rsiScopeTeal}
\addplot[rsi scoped line] table[x=round,y=scopedMean] \rsiFixed;
\addlegendentry{Scoped retrieval}
\addplot[rsi global line] table[x=round,y=globalMean] \rsiFixed;
\addlegendentry{Global-ORC}
\addplot[rsi static line] table[x=round,y=staticMean] \rsiFixed;
\addlegendentry{Static = Batch-ORC}

\nextgroupplot[
  title={(b) Permissive baselines},
]
\rsitrajectoryband{\rsiFixed}{round}{static}{rsiStaticGray}
\rsitrajectoryband{\rsiFixed}{round}{latest}{rsiLatestOrange}
\rsitrajectoryband{\rsiFixed}{round}{replay}{rsiGlobalNavy}
\addplot[rsi latest line] table[x=round,y=latestMean] \rsiFixed;
\addlegendentry{Latest-only}
\addplot[rsi replay line] table[x=round,y=replayMean] \rsiFixed;
\addlegendentry{Replay}
\addplot[rsi static line] table[x=round,y=staticMean] \rsiFixed;
\addlegendentry{Static}
\end{groupplot}
\end{tikzpicture}
\endgroup

%% file: generated_external.tex
\begin{table}[t]
\centering
\small
\caption{Terminal-policy transfer. Values are means with seed-bootstrap 95\% intervals. HumanEval+ uses the fixed 32-task sample; GPT-OSS uses each stream's 18-task sealed final bank. $^\dagger$Task-specific retrieval over the same archived completions.}
\label{tab:transfer}
\begin{tabular}{lcc}
\toprule
Method & HumanEval+ pass rate & GPT-OSS hidden pass rate \\
\midrule
Static & 0.914 [0.898, 0.930] & 0.941 [0.907, 0.974] \\
Latest-only & 0.918 [0.891, 0.941] & 0.911 [0.886, 0.937] \\
Replay & 0.922 [0.895, 0.945] & 0.916 [0.878, 0.951] \\
ORC & 0.895 [0.867, 0.922] & 0.951 [0.926, 0.968] \\
Batch-ORC & 0.914 [0.898, 0.930] & 0.941 [0.907, 0.974] \\
Task-specific retrieval$^{\dagger}$ & --- & 0.960 [0.930, 0.987] \\
\bottomrule
\end{tabular}
\end{table}

%% file: generated_randomized_entry.tex
\begin{table}[t]
\centering
\small
\setlength{\tabcolsep}{5pt}
\caption{Balanced randomized-entry replication (18 streams; two per first family). Brackets show stream-bootstrap 95\% intervals.}
\label{tab:randomized-entry}
\begin{tabular}{lcccc}
\toprule
Deployment & Accepted & Harmful & Checkpoint $t{=}1$ & Non-current $\Delta$ \\
\midrule
Global ORC & 5 & 3 & 0.761 [0.709, 0.802] & -0.047 [-0.120, 0.005] \\
Scoped ORC & 5 & 0 & 0.802 [0.779, 0.826] & 0.000 [0.000, 0.000] \\
\bottomrule
\end{tabular}
\end{table}

%% file: generated_reviewer_online.tex
\begin{table}[t]
\centering
\caption{Online controls on 27 paired 12-round streams. Intervals are computed by resampling whole
streams. Harm reports accepted steps that lower the next checkpoint score and the number
of streams affected. Cross-family drops count score decreases on other families after
deployment, potentially several per step. Scoped-Cumulative may install several skills in one step.}
\label{tab:reviewer-online}
\resizebox{\linewidth}{!}{%
\begin{tabular}{lrrrrrr}
\toprule
Method & Trajectory & Final checkpoint & Final bank & Accepted & Harm (streams) & Cross-family drops \\
\midrule
Scoped-Replay & 0.884 [0.867, 0.901] & 0.961 [0.947, 0.975] & 0.953 [0.941, 0.965] & 287 & 11 (11/27) & 0 \\
Matched Scoped-ORC & 0.848 [0.825, 0.871] & 0.900 [0.871, 0.927] & 0.862 [0.829, 0.893] & 63 & 0 (0/27) & 0 \\
Multiversion-ORC & 0.822 [0.798, 0.846] & 0.849 [0.820, 0.878] & 0.801 [0.772, 0.830] & 49 & 8 (7/27) & 20 \\
Scoped-Learned & 0.810 [0.791, 0.828] & 0.833 [0.809, 0.857] & 0.805 [0.777, 0.832] & 46 & 8 (8/27) & 14 \\
Scoped-Cumulative & 0.798 [0.772, 0.822] & 0.849 [0.814, 0.883] & 0.820 [0.788, 0.854] & 34 & 0 (0/27) & 0 \\
Global-Private & 0.788 [0.758, 0.814] & 0.792 [0.761, 0.819] & 0.746 [0.712, 0.777] & 12 & 5 (5/27) & 31 \\
Global-Latest & 0.696 [0.681, 0.711] & 0.666 [0.626, 0.704] & 0.678 [0.643, 0.714] & 324 & 168 (27/27) & 602 \\
Global-Replay & 0.744 [0.722, 0.765] & 0.765 [0.741, 0.788] & 0.746 [0.729, 0.763] & 151 & 71 (27/27) & 236 \\
\bottomrule
\end{tabular}}
\end{table}

\begin{table}[t]
\centering
\caption{Online resource use. Logged calls and token totals include cached responses used by the
completed runs. Billed completions and charges cover all attempts, including retries. Ambiguous
requests timed out after dispatch with unresolved billing status and are charged conservatively.}
\label{tab:reviewer-cost}
\resizebox{\linewidth}{!}{%
\begin{tabular}{lrrrrrr}
\toprule
Method & Logged calls & Input tokens & Output tokens & Billed completions & Ambiguous & Charge \\
\midrule
Scoped-Replay & 21,945 & 13,485,644 & 1,607,113 & 643 & 0 & \$0.073 \\
Matched Scoped-ORC & 22,113 & 13,570,048 & 1,830,822 & 0 & 0 & \$0.000 \\
Multiversion-ORC & 22,113 & 13,560,822 & 2,147,753 & 1,048 & 0 & \$0.227 \\
Scoped-Learned & 22,113 & 13,571,721 & 2,117,597 & 139 & 0 & \$0.053 \\
Scoped-Cumulative & 28,512 & 15,389,264 & 2,385,540 & 797 & 0 & \$0.220 \\
Global-Private & 22,089 & 13,549,759 & 2,584,241 & 447 & 0 & \$0.082 \\
Global-Latest & 5,265 & 6,324,624 & 985,644 & 4,728 & 4 & \$1.420 \\
Global-Replay & 22,065 & 13,472,016 & 3,054,631 & 12,612 & 13 & \$2.596 \\
\bottomrule
\end{tabular}}
\end{table}

%% file: generated_second_backbone.tex
\begin{table}[htbp]
\centering
\small
\caption{Full GPT-OSS-120B reruns on the same 27 twelve-round streams. Brackets are whole-stream
bootstrap 95\% intervals. Harm reports accepted steps with lower next-checkpoint utility and
the number of affected streams. Cross-family drops count noncurrent-family decreases after
accepted updates.}
\label{tab:second-backbone}
\resizebox{\linewidth}{!}{%
\begin{tabular}{lrrrrrr}
\toprule
Condition & Trajectory & Final checkpoint & Final bank & Accepts & Harm (streams) & Cross-family drops \\
\midrule
Static & 0.909 [0.884, 0.933] & 0.909 [0.884, 0.933] & 0.941 [0.921, 0.959] & 0 & 0 (0/27) & 0 \\
Global-ORC & 0.914 [0.890, 0.938] & 0.916 [0.891, 0.940] & 0.939 [0.919, 0.957] & 1 & 0 (0/27) & 0 \\
Scoped-ORC & 0.914 [0.890, 0.937] & 0.917 [0.892, 0.941] & 0.943 [0.923, 0.963] & 2 & 0 (0/27) & 0 \\
Multiversion-ORC & 0.916 [0.892, 0.940] & 0.920 [0.895, 0.944] & 0.943 [0.922, 0.962] & 2 & 0 (0/27) & 0 \\
\bottomrule
\end{tabular}}
\end{table}

\begin{table}[htbp]
\centering
\small
\caption{Paired GPT-OSS-120B effects with Scoped-ORC as the minuend. Holm correction covers
the three trajectory contrasts. Terminal and final-bank intervals are descriptive.}
\label{tab:second-backbone-effects}
\resizebox{\linewidth}{!}{%
\begin{tabular}{lrrrr}
\toprule
Comparator & Trajectory & Final checkpoint & Final bank & Holm $p$ \\
\midrule
Global-ORC & -0.0002 [-0.0071, 0.0066] & +0.0010 [-0.0093, 0.0123] & +0.0045 [0.0000, 0.0116] & 1 \\
Multiversion-ORC & -0.0024 [-0.0071, 0.0000] & -0.0031 [-0.0093, 0.0000] & +0.0005 [0.0000, 0.0015] & 1 \\
Static & +0.0052 [0.0000, 0.0133] & +0.0080 [0.0000, 0.0201] & +0.0027 [0.0000, 0.0073] & 1 \\
\bottomrule
\end{tabular}}
\end{table}

%% file: references.bib
@misc{asadolahi2026memory,
  author = {Mohammad Asadolahi and Amir Amini and Samira Talebi and Amirfarhad Farhadi and Azadeh Zamanifar},
  title = {Memory Reward Inflation in Self-Improving LLM Agents},
  year = {2026},
  note = {arXiv preprint arXiv:2608.00017}
}

@misc{herel2024collapse,
  author = {David Herel and Tomas Mikolov},
  title = {Collapse of Self-Trained Language Models},
  year = {2024},
  note = {arXiv preprint arXiv:2404.02305}
}

@inproceedings{hu2024adas,
  author = {Shengran Hu and Cong Lu and Jeff Clune},
  title = {Automated Design of Agentic Systems},
  booktitle = {International Conference on Learning Representations},
  year = {2025},
  url = {https://arxiv.org/abs/2408.08435}
}

@inproceedings{huang2023cannot,
  author = {Jie Huang and Xinyun Chen and Swaroop Mishra and Huaixiu Steven Zheng and Adams Wei Yu and Xinying Song and Denny Zhou},
  title = {Large Language Models Cannot Self-Correct Reasoning Yet},
  booktitle = {International Conference on Learning Representations},
  year = {2024},
  url = {https://arxiv.org/abs/2310.01798}
}

@misc{iacob2026redqueen,
  author = {Alex Iacob and Andrej Jovanovi{\'c} and William F. Shen and Daniel Burkhardt and Meghdad Kurmanji and Nurbek Tastan and Lorenzo Sani and Niccol{\`o} Alberto Elia Venanzi and Ambroise Odonnat and Zeyu Cao and others},
  title = {The Red Queen G{\"o}del Machine: Co-Evolving Agents and Their Evaluators},
  year = {2026},
  note = {arXiv preprint arXiv:2606.26294}
}

@misc{liu2026mendel,
  author = {Changzhi Liu and Yilun Liu and Sikuan Yan and Volker Tresp and Yunpu Ma},
  title = {Mendel G{\"o}del Machine: Recursive Self-Improving Coding Agents via Comparative Evolution},
  year = {2026},
  note = {arXiv preprint arXiv:2608.07645}
}

@inproceedings{liu2023evalplus,
  author = {Jiawei Liu and Chunqiu Steven Xia and Yuyao Wang and Lingming Zhang},
  title = {Is Your Code Generated by {ChatGPT} Really Correct? Rigorous Evaluation of Large Language Models for Code Generation},
  booktitle = {Advances in Neural Information Processing Systems},
  year = {2023},
  url = {https://openreview.net/forum?id=1qvx610Cu7}
}

@inproceedings{madaan2023selfrefine,
  author = {Aman Madaan and Niket Tandon and Prakhar Gupta and Skyler Hallinan and Luyu Gao and Sarah Wiegreffe and Uri Alon and Nouha Dziri and Shrimai Prabhumoye and Yiming Yang and others},
  title = {Self-Refine: Iterative Refinement with Self-Feedback},
  booktitle = {Advances in Neural Information Processing Systems},
  year = {2023},
  url = {https://arxiv.org/abs/2303.17651}
}

@misc{pan2024rewardhacking,
  author = {Jane Pan and He He and Samuel R. Bowman and Shi Feng},
  title = {Spontaneous Reward Hacking in Iterative Self-Refinement},
  year = {2024},
  note = {arXiv preprint arXiv:2407.04549}
}

@misc{shafayat2025selftrain,
  author = {Sheikh Shafayat and Fahim Tajwar and Ruslan Salakhutdinov and Jeff Schneider and Andrea Zanette},
  title = {Can Large Reasoning Models Self-Train?},
  year = {2025},
  note = {arXiv preprint arXiv:2505.21444}
}

@inproceedings{shinn2023reflexion,
  author = {Noah Shinn and Federico Cassano and Edward Berman and Ashwin Gopinath and Karthik Narasimhan and Shunyu Yao},
  title = {Reflexion: Language Agents with Verbal Reinforcement Learning},
  booktitle = {Advances in Neural Information Processing Systems},
  year = {2023},
  url = {https://arxiv.org/abs/2303.11366}
}

@misc{singh2023restem,
  author = {Avi Singh and John D. Co-Reyes and Rishabh Agarwal and Ankesh Anand and Piyush Patil and Xavier Garcia and Peter J. Liu and James Harrison and Jaehoon Lee and Kelvin Xu and others},
  title = {Beyond Human Data: Scaling Self-Training for Problem-Solving with Language Models},
  year = {2023},
  note = {arXiv preprint arXiv:2312.06585}
}

@misc{song2024mindgap,
  author = {Yuda Song and Hanlin Zhang and Carson Eisenach and Sham Kakade and Dean Foster and Yudong Ghai},
  title = {Mind the Gap: Examining the Self-Improvement Capabilities of Large Language Models},
  year = {2024},
  note = {arXiv preprint arXiv:2412.02674}
}

@misc{wang2026compound,
  author = {Wenxiao Wang and Priyatham Kattakinda and Soheil Feizi},
  title = {Do Agent Optimizers Compound? A Continual-Learning Evaluation on Terminal-Bench 2.0},
  year = {2026},
  note = {arXiv preprint arXiv:2607.14004}
}

@misc{wang2026metaskill,
  author = {Zefeng Wang and Minxi Yan and Jinhe Bi and Sikuan Yan and Volker Tresp and Yunpu Ma},
  title = {MetaSkill-Evolve: Recursive Self-Improvement of LLM Agents via Two-Timescale Meta-Skill Evolution},
  year = {2026},
  note = {arXiv preprint arXiv:2607.05297}
}

@misc{wu2024metarewarding,
  author = {Tianhao Wu and Weizhe Yuan and Olga Golovneva and Jing Xu and Yuandong Tian and Jiantao Jiao and Jason Weston and Sainbayar Sukhbaatar},
  title = {Meta-Rewarding Language Models: Self-Improving Alignment with LLM-as-a-Meta-Judge},
  year = {2024},
  note = {arXiv preprint arXiv:2407.19594}
}

@misc{yin2024godelagent,
  author = {Xunjian Yin and Xinyi Wang and Liangming Pan and Li Lin and Xiaojun Wan and William Yang Wang},
  title = {G{\"o}del Agent: A Self-Referential Agent Framework for Recursive Self-Improvement},
  year = {2024},
  note = {arXiv preprint arXiv:2410.04444}
}

@misc{yuan2024selfrewarding,
  author = {Weizhe Yuan and Richard Yuanzhe Pang and Kyunghyun Cho and Xian Li and Sainbayar Sukhbaatar and Jing Xu and Jason Weston},
  title = {Self-Rewarding Language Models},
  year = {2024},
  note = {arXiv preprint arXiv:2401.10020}
}

@inproceedings{zelikman2022star,
  author = {Eric Zelikman and Yuhuai Wu and Jesse Mu and Noah D. Goodman},
  title = {{STaR}: Bootstrapping Reasoning with Reasoning},
  booktitle = {Advances in Neural Information Processing Systems},
  year = {2022},
  url = {https://arxiv.org/abs/2203.14465}
}

@misc{zhang2025dgm,
  author = {Jenny Zhang and Shengran Hu and Cong Lu and Robert Lange and Jeff Clune},
  title = {Darwin G{\"o}del Machine: Open-Ended Evolution of Self-Improving Agents},
  year = {2025},
  note = {arXiv preprint arXiv:2505.22954}
}

@inproceedings{zhuo2024bigcodebench,
  author = {Terry Yue Zhuo and Minh Chien Vu and Jenny Chim and Han Hu and Wenhao Yu and Ratnadira Widyasari and Imam Nur Bani Yusuf and Haolan Zhan and Junda He and Indraneil Paul and others},
  title = {BigCodeBench: Benchmarking Code Generation with Diverse Function Calls and Complex Instructions},
  booktitle = {International Conference on Learning Representations},
  year = {2025},
  url = {https://arxiv.org/abs/2406.15877}
}

@inproceedings{ma2026benchmarking,
  title={Benchmarking Continual Agent Memory for Online Learning, Transfer, and Forgetting},
  author={Ma, Zihang and Liu, Jinyi and Tang, Hongyao and Ma, Yi and Wang, Ruitao and Yuan, Yifu and Zheng, Yan and Hao, Jianye},
  booktitle={ICLR Workshop on Lifelong Agents: Learning, Aligning, Evolving},
  year={2026},
  url={https://openreview.net/forum?id=MSXbrNExax}
}

@inproceedings{zheng2024synapse,
  title = {{Synapse}: Trajectory-as-Exemplar Prompting with Memory for Computer Control},
  author = {Zheng, Longtao and Wang, Rundong and Wang, Xinrun and An, Bo},
  booktitle = {International Conference on Learning Representations},
  year = {2024},
  url = {https://openreview.net/forum?id=Pc8AU1aF5e}
}

@inproceedings{wang2025agentworkflow,
  title = {Agent Workflow Memory},
  author = {Wang, Zora Zhiruo and Mao, Jiayuan and Fried, Daniel and Neubig, Graham},
  booktitle = {Proceedings of the 42nd International Conference on Machine Learning},
  series = {Proceedings of Machine Learning Research},
  volume = {267},
  pages = {63897--63911},
  year = {2025},
  publisher = {PMLR},
  url = {https://proceedings.mlr.press/v267/wang25bx.html}
}

@misc{openai2025gptoss,
  title = {{gpt-oss-120b} \& {gpt-oss-20b} Model Card},
  author = {{OpenAI}},
  year = {2025},
  note = {arXiv:2508.10925},
  url = {https://arxiv.org/abs/2508.10925}
}
